\documentclass[onecolumn]{article} 
\usepackage[top=3.6cm, bottom=3.2cm, left=2.5cm, right=2.5cm]{geometry}
\usepackage[colorlinks=true,linkcolor=red,citecolor=blue]{hyperref}
\usepackage{amsthm,amsmath,amssymb}
\usepackage{mathrsfs}
\usepackage{upgreek}

\usepackage{tabularx}
\usepackage{graphicx}
\usepackage{subfigure}

\usepackage{cite}
\usepackage{ulem}
\usepackage{url}

\usepackage{xurl}
\usepackage{multirow}
\usepackage{xcolor}

\begin{document}
\title{Invariant Features for Global Crop Type Classification}
\author{
Xin-Yi Tong$^{1}$, Sherrie Wang$^{1,2,3}$ 
\vspace{3mm}
\\
$^1$ \small\textit{Laboratory for Information and Decision Systems, MIT} \\
$^2$ \small\textit{Department of Mechanical Engineering, MIT} \\
$^3$ \small\textit{Institute for Data, Systems, and Society, MIT} \\
\small\textit{Email: \{xytong, sherwang\}@mit.edu}
}
\date{}
\maketitle

\begin{abstract}
Accurate global crop type mapping supports agricultural monitoring and food security, yet remains limited by the scarcity of labeled data in many regions. A key challenge is enabling models trained in one geography to generalize reliably to others despite shifts in climate, phenology, and spectral characteristics.
In this work, we show that geographic transfer in crop classification is primarily governed by the ability to learn invariant structure in multispectral time series. To systematically study this, we introduce CropGlobe, a globally distributed benchmark dataset of 300,000 samples spanning eight countries and five continents, and define progressively harder transfer settings from cross-country to cross-hemisphere.
Across all settings, we find that simple spectral--temporal representations outperform both handcrafted features and modern geospatial foundation model embeddings. We propose CropNet, a lightweight convolutional architecture that jointly convolves across spectral and temporal dimensions to learn invariant crop signatures. Despite its simplicity, CropNet consistently outperforms larger transformer-based and foundation-model approaches under geographic domain shift.
To further improve robustness to geographic variation, we introduce augmentations that simulate shifts in crop phenology and reflectance. Combined with CropNet, this yields substantial gains under large domain shifts. Our results demonstrate that inductive bias toward joint spectral--temporal structure is more critical for transfer than model scale or pretraining, pointing toward a scalable and data-efficient paradigm for worldwide agricultural mapping. Our data and code are available publicly at \url{https://github.com/x-ytong/CropNet/}.
\end{abstract}

\section{Introduction}
Accurately identifying crop types and their spatial distribution at the global scale is essential for ensuring food security, guiding agricultural production, and informing agricultural policy \cite{importance1, importance2, importance3}. As pressures from climate change, population growth, and increasing food demand continue to intensify, dynamic monitoring of agricultural systems has become a central component of global sustainable development \cite{importance4, importance5}.

Traditionally, crop type information has been obtained through field surveys and agricultural censuses. While reliable, these approaches are labor-intensive and costly, making it difficult to update agricultural statistics in a timely manner \cite{challenge1}. With the rapid development of Earth observation technologies and machine learning methods, remote sensing has played an increasingly prominent role in agricultural monitoring. However, these products are largely limited to countries or regions with strong statistical capacities \cite{usda_cdl, aafc_aci, uk_crome} and such reliable crop type information remains unavailable across large parts of the world. 
Recent global efforts such as WorldCereal \cite{worldcereal} represent important progress, but remain limited to binary or aggregated crop classes, leaving global multi-crop mapping an open challenge.

The fundamental bottleneck in expanding crop type maps is the lack of ground reference data. Without labeled samples in the target region, machine learning models often suffer substantial performance degradation when applied to unseen areas \cite{challenge2, challenge3}. This raises a critical research question: what representations can transfer across geographies without local supervision? 

We frame this as a problem of invariance. If crop types exhibit sufficient commonality across regions, then transferable models must rely on features that are stable under shifts in climate, phenology, and background conditions, rather than region-specific correlations. Identifying such invariant features is key to enabling cross-region crop type classification without ground labels. While a wide range of methods have been proposed for crop mapping --- from harmonic features to one-dimensional convolutional neural networks (1D CNNs) to modern geospatial foundation models (GFMs) --- their relative performance in cross-region transfer has not been systematically compared at a global scale.

In this paper, we show that geographically transferable crop signatures reside in the joint spectral--temporal space, and identifying these invariant features can be achieved with a specific inductive bias: CNNs with kernels that convolve jointly across both dimensions to capture coordinated spectral--temporal trajectories. The central challenge for learning invariant features is not model capacity (i.e., parameter count), but how the spectral and temporal dimensions are represented. 

To systematically study feature invariance, we construct CropGlobe, a dataset of 300,000 crop type samples across eight countries and five continents. We use this benchmark to evaluate feature transferability across three progressively more difficult scenarios: cross-country, cross-continent, and cross-hemisphere classification. We then propose CropNet, a lightweight convolutional architecture designed to enforce this spectral--temporal inductive bias. By convolving jointly across the temporal and spectral dimensions, CropNet is able to extract representations that are significantly more robust to geographic domain shifts than prior 1D temporal or foundation model baselines. We further complement this architecture with a tailored data augmentation strategy that simulates realistic phenological and spectral shifts, including time-warping and reflectance magnitude variations.

Our contributions are summarized as follows:

\begin{itemize}
    \item \textbf{CropGlobe Benchmark:} We construct a multi-continental dataset spanning eight countries and six key crops, enabling systematic evaluation of cross-region generalization.

    \item \textbf{Analysis of Feature Invariance:} We conduct a comprehensive comparison of representations --- including harmonic features, various deep learning models, and GFM embeddings --- under progressively difficult geographic transfer settings.

    \item \textbf{Spectral--Temporal Inductive Bias (CropNet):} We propose CropNet, an architecture that leverages joint spectral--temporal convolutions to extract invariant features. We show that this inductive bias outperforms both traditional temporal baselines and GFMs, with an additional 7\% accuracy gain from targeted data augmentation under limited training diversity.
\end{itemize}

\section{Related Work}
Previous studies devoted considerable effort to cross-geographic crop classification. Among them, some focused on constructing or learning more transferable remote sensing features, while others sought to reduce representation mismatch caused by phenological shifts.

\textbf{Harmonic Features:} Harmonic regression captures the within-year phenological rhythm of crops using a low-dimensional set of parameters \cite{harmonic1, harmonic2}. As a result, it is generally more robust to missing observations, noise, and differences in observation frequency. In the absence of field-level labels, harmonic features were used for cross-region crop classification in the U.S. Midwest \cite{challenge1}, where a random forest (RF) classifier trained in one state was transferred to other states or other years. This demonstrated that, even without ground observations in the target region, models based on phenological features can still retain a certain degree of transferability. However, what such methods transfer is essentially the relative temporal structure of crop phenology. Once there are substantial differences between the source and target regions in planting dates, growth rates, or sensor background conditions, the stability of harmonic parameters declines.

\textbf{Temporal Median Features:} Median features aggregate multi-temporal observations within predefined temporal windows and use the median to represent the spectral state at each stage \cite{median1, median2}, thereby reducing the influence of clouds, noise, and anomalous observations. Biweekly Sentinel-2 (S2) median features were used for global cross-region soybean mapping \cite{medianTransfer}. By aligning time series based on crop phenology, consistency between regions was improved, which in turn enhanced cross-region transferability. This suggests that, although median features are simple in form, they can still achieve strong generalization when combined with appropriate temporal alignment. However, when features from the Northern Hemisphere were used to classify soybean in the Southern Hemisphere, performance dropped sharply compared with within-hemisphere transfer. This may be because the combination of median features and phenological alignment captures only the overall seasonal trend and still cannot fully resolve cross-region representation mismatch.

\textbf{GFM Embeddings:} 
Handcrafted features often rely on task-specific design and may not generalize well across regions. In contrast, GFMs learn representations through large-scale self-supervised pretraining on diverse remote sensing data and provide general-purpose embeddings for downstream tasks. However, whether such pretraining promotes geographic invariance or instead encodes region-specific signals (e.g., climate, topography, or phenology) remains an open question.

Among them, Presto \cite{presto} is a lightweight pretrained Transformer designed for pixel-level remote sensing time series. It learns representations in a self-supervised manner by reconstructing masked time steps and sensor modalities, with an emphasis on unified modeling of the temporal dimension, multi-sensor inputs, and missing observations. By contrast, AlphaEarth \cite{Alphaearth} is built around organizing multi-source Earth observation data into a globally consistent embedding field, that is, a general-purpose embedding representation that can be directly queried for each location on the Earth’s surface, thereby supporting large-scale mapping and monitoring under limited-label settings.

However, the requirements of agricultural applications are not entirely the same as those of general land-surface tasks. Results from yield prediction, tillage mapping, and cover crop mapping \cite{harvesting} show that AlphaEarth embeddings are competitive in agricultural downstream tasks, but also indicate limitations in temporal sensitivity and spatial transfer stability. For cross-domain crop classification, although GFM embeddings provide strong general-purpose priors, their learned representations are not necessarily naturally aligned to crop phenology, and their effectiveness still needs to be validated in crop type mapping scenarios.

\textbf{Data Augmentation for Transferability:} Regardless of the feature type, failure in cross-region crop classification often lies in the inability of the training data to cover phenological variation in the target domain. Accordingly, an effective way to mitigate this problem is to actively simulate possible domain shifts, so that the model can encounter and adapt to such variation during training.

Temporal encoding \cite{Temporally} was proposed for cross-year crop classification transfer. Specifically, it uses Random Observations Selection to simulate changes in interannual observation sparsity, and Random Day Shifting to simulate phenological shifts between years, thereby improving robustness to temporal changes. For cross-region crop classification transfer, TimeMatch \cite{TimeMatch} estimates the temporal shift between the source and target domains by comparing model predictions under different temporal offsets, and then achieves alignment by iteratively updating the model with pseudo-label-based self-training. These studies show that introducing temporal variation can effectively improve transferability in crop classification.

However, these methods mainly focus on “early-or-late” shifts. In reality, cross-region differences are often more complex, the same crop may be planted earlier or later, but it may also grow faster or slower, with its growth rhythm stretched or compressed. More importantly, cross-region differences are not only about timing. Soil background, management practices, and climatic conditions can also lead to systematic shifts in overall reflectance levels. As a result, time shift alone is still insufficient to capture the full range of cross-region variation.

\section{CropGlobe Dataset}
To facilitate our investigation into feature invariance in global crop type classification, we construct the CropGlobe dataset using 2023 as the reference year. It is built upon publicly available crop type reference products collected from eight countries across five continents: Argentina (ARG), Australia (AUS), Belgium (BEL), China (CHN), France (FRA), the Netherlands (NLD), the United Kingdom (GBR), and the United States (USA). To the best of our knowledge, these products represent the high-quality public crop maps available to us for this study.

\begin{figure*}[htb!]
\centering
\includegraphics[width=1\textwidth]
{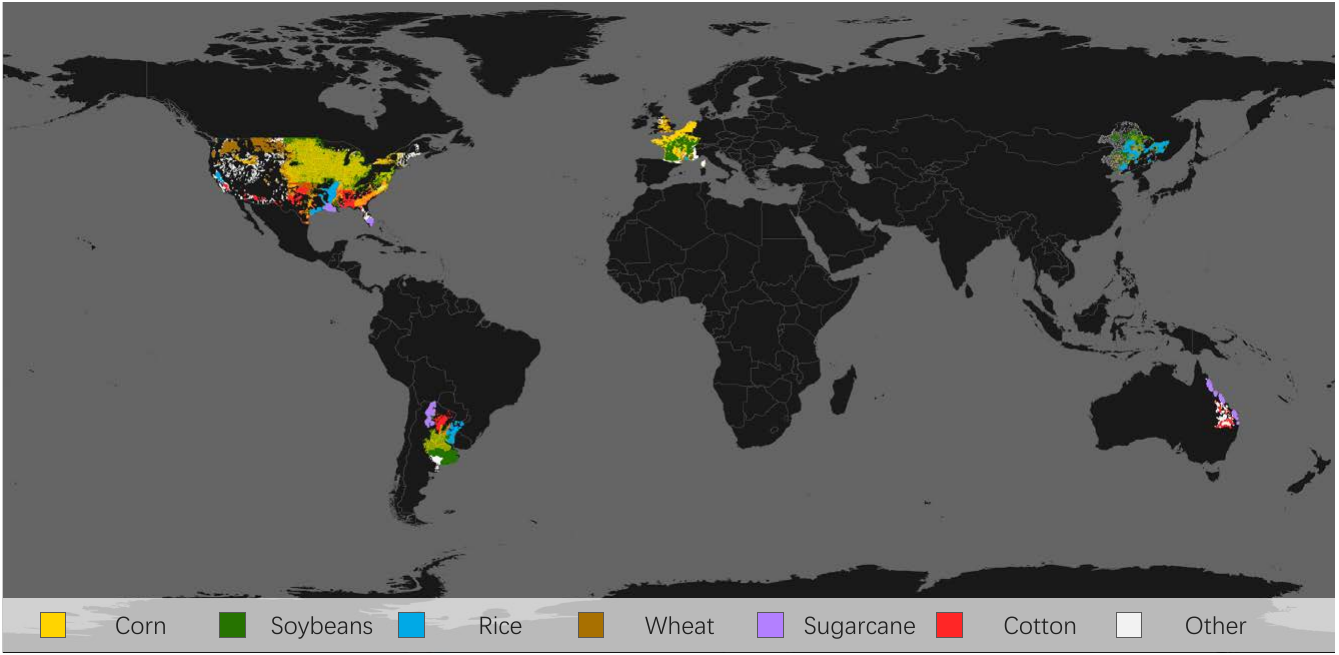}
\caption{The geographical distribution and category system of the CropGlobe dataset. It contains 300,000 pixel-level samples from eight countries across five continents: ARG, AUS, BEL, CHN, FRA, GBR, NLD, and USA. For some countries, the sample coverage is not national in extent because the original reference crop type products are region-specific rather than country-wide.}
\label{figure:dataglobe}
\end{figure*}

We focus on six globally significant food and industrial crops: corn, soybeans, rice, wheat, sugarcane, and cotton, whose geographical distribution is displayed in Fig.~\ref{figure:dataglobe}. Including an ``Other'' class, CropGlobe consists of seven crop categories and contains more than 300,000 samples, each corresponding to a georeferenced pixel-level location on the Earth’s surface.

The following sections provide descriptions of crop type products (Section \ref{sec:referencedata}), satellite data source (Section \ref{sec:satellitedata}), sample extraction procedures (Section \ref{sec:samplecollection}), and crop phenologies by region (Section \ref{sec:datavisualization}).

\subsection{Reference Data}
\label{sec:referencedata}
Reference data are compiled from publicly available crop type mapping products selected for their data quality, public accessibility, and geographic diversity. Specifically, we use Argentina National Map of Crops (ANMC) \cite{ANMC} for ARG, Queensland Seasonal Crop (QSC) \cite{QSC} for AUS, Landbouwgebruikspercelen (LGP) \cite{LGP} for BEL, Crop Type in Northeast China (CTNC) \cite{CTNC} for CHN, Registre Parcellaire Graphique (RPG) \cite{RPG} for FRA, Crop Map of England (CROME) \cite{uk_crome} for GBR, Basisregistratie Gewaspercelen (BRP) \cite{BRP} for NLD, and Cropland Data Layer (CDL) \cite{usda_cdl} for USA. 

\begin{table*}[t]
\centering
\caption{Reference crop type products used in CropGlobe. All products correspond to the year 2023. ``Classes'' denotes the number of crop categories in the original product, while ``Classes Used'' denotes the categories retained in this study. ``Selected target crops'' indicates which of the six target crops are available in each product, depending on product coverage.}
\vspace{3mm}
\resizebox{0.8\textwidth}{16mm}{
\begin{tabular}{llllll}
\hline
\textbf{Country} & \textbf{Product} & \textbf{Format} & \textbf{Label Source} & \textbf{Classes} & \textbf{Classes Used} \\
\hline
ARG & ANMC  & raster / 30 m   & survey + classification & 15  & Selected target crops + Other \\
AUS & QSC   & polygon vector  & survey + classification & 3   & Cotton, Sugarcane + Other     \\
BEL & LGP   & polygon vector  & farmer-reported         & 297 & Selected target crops + Other \\
CHN & CTNC  & raster / 10 m   & survey + classification & 3   & Rice, Corn, Soybeans + Other  \\
FRA & RPG   & polygon vector  & farmer-reported         & 372 & Selected target crops + Other \\
GBR & CROME & polygon vector  & survey + classification & 81  & Selected target crops + Other \\
NLD & BRP   & polygon vector  & farmer-reported         & 371 & Selected target crops + Other \\
USA & CDL   & raster / 30 m   & survey + classification & 254 & Selected target crops + Other \\
\hline
\end{tabular}}
\label{table:referencedata}
\end{table*}

All reference products correspond to the year 2023. Where reported by the original products, mapping accuracy is generally high, with user and producer accuracies of around 95\% and 90\% for QSC \cite{queensland}, producer accuracy above 92\% for CTNC \cite{CTNC}, user and producer accuracies above 83\% for major classes in CROME \cite{uk_crome}, and 85-95\% accuracy for major classes in CDL \cite{usda_cdl}. Key characteristics of these reference products are summarized in Table~\ref{table:referencedata}, and direct access links are provided in Appendix \ref{sec:appendix_links}.

It should be noted that some products do not cover the full extent of the corresponding country. For example, QSC covers only Queensland in Australia, and CTNC covers only Northeast China. Nevertheless, for simplicity, we refer to these products by country names throughout the paper. In addition, the classes available in each product are constrained by the original mapping scheme rather than by the agricultural importance of particular crops in that country. For instance, although wheat is one of the major crops in Australia, it is not included in this study because the QSC winter-season product provides only a generic ``Crop'' class rather than crop-specific labels.

To ensure label quality and maintain a consistent category system, we perform category extraction and merging. For BEL, FRA, and NLD, which are based on high-accuracy farmer declarations, we remove non-agricultural classes and merge crop types outside the six target crops into a single ``Other'' class.

For ARG, GBR and USA, where the maps are derived from automatic classification with class-wise accuracy metadata, we apply stricter filtering. In ANMC and CROME, classes with user accuracy below 80\% are removed (all six target classes exceed this threshold), and the remaining non-target classes are merged into ``Other''. In CDL, accuracy is reported at the state level; for each target class, only states with both producer and user accuracy above 85\% are retained. Within those states, non-agricultural classes and double-cropping classes are removed, and only non-target classes with accuracy above 80\% are retained and merged into ``Other''. The filtering criteria differ between ARG, GBR and USA because ANMC and CROME have lower overall accuracy than CDL, and equally strict filtering in ARG and GBR would remove nearly all non-target classes and leave too few samples for ``Other''. 

\begin{sloppypar}
In AUS and CHN, the source maps are three-class products, so no further class selection or merging is required. However, these maps do not provide explicit labels for non-target crop categories. To capture them, we incorporate ESA WorldCover \cite{WorldCover}; specifically, pixels in the GEE collection \texttt{ESA/WorldCover/v200} labeled as ``cropland'' but not recognized by the source maps are assigned to the ``Other'' class.
\end{sloppypar}

\subsection{Satellite Data}
\label{sec:satellitedata}
Sentinel-2 is part of the Copernicus Programme led by the European Space Agency (ESA), with a mission focused on the dynamic monitoring of terrestrial surfaces and vegetation. 
The system consists of two satellites equipped with the Multi-Spectral Instrument (MSI), which acquires imagery across 13 spectral bands spanning the visible, near-infrared, and shortwave infrared regions at spatial resolutions of 10--60 meters.

Combined, the two satellites provide a global revisit frequency of five days. With its relatively high spatial resolution, S2 is well suited for multispectral time-series analysis to capture phenological patterns and differentiate crop types. It has become one of the most widely used data sources for agricultural dataset development and global crop mapping, such as in CropHarvest \cite{cropharvest} and WorldCereal \cite{worldcereal}.

\begin{sloppypar}
We obtain S2 Level-2A imagery from the Google Earth Engine (GEE) collection \texttt{COPERNICUS/S2\_SR\_HARMONIZED}, which provides surface reflectance corrected for atmospheric effects. Cloudy pixels are filtered using Cloud Score+ from the GEE collection \texttt{GOOGLE/CLOUD\_SCORE\_PLUS/V1/S2\_HARMONIZED}, with a threshold of 0.75, such that only pixels with non-cloud probability $\geq 75\%$ are retained. 
Of the 13 available S2 bands, we exclude bands B1, B9, and B10 (coastal aerosol, water vapor, and cirrus) and retain ten bands relevant for crop type classification (B2, B3, B4, B8, B5, B6, B7, B8A, B11, and B12). 

For countries in the Northern Hemisphere, we collect S2 from January 1 to December 31, 2023. For countries in the Southern Hemisphere, the collection period is shifted by six months, from July 1, 2022 to June 30, 2023, to account for the phenological offset between hemispheres.
\end{sloppypar}

\begin{figure*}[htb!]
\centering
\includegraphics[width=0.75\textwidth]
{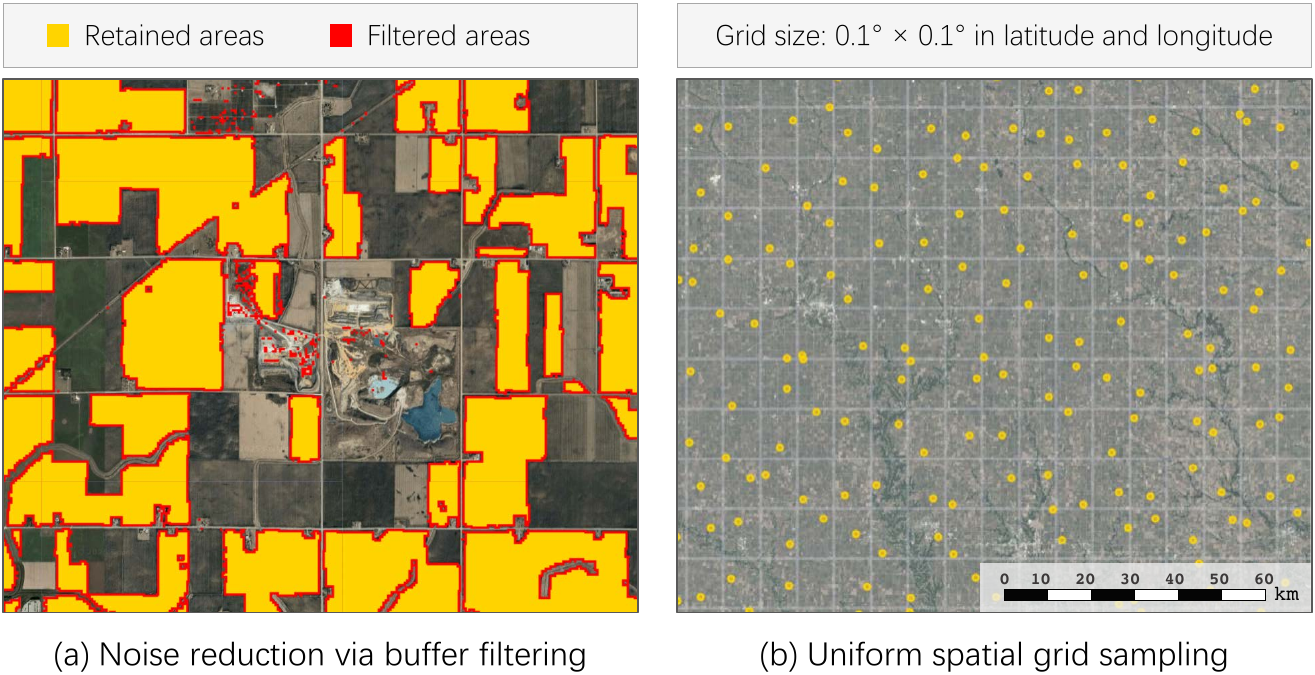}
\caption{Illustration of noise reduction via buffered filtering and sample decorrelation through grid-based sampling.}
\label{figure:databuffer}
\end{figure*}

\begin{table*}[htb!]
\centering
\caption{Class distribution of the CropGlobe dataset by country. Values marked with $^*$ correspond to samples derived from crop maps with ternary labels. The ``Other'' class may contain corn, soybeans, rice, wheat, sugarcane, or cotton, see Section \ref{sec:setup} for processing details.}
\vspace{3mm}
\resizebox{0.75\textwidth}{15mm}{
\begin{tabular}{lcrcrcrcrrcrcrcr}
\hline
\textbf{Country} & & \textbf{Corn} & & \textbf{Soybeans} & & \textbf{Rice} & & \textbf{Wheat} & \textbf{Sugarcane} & & \textbf{Cotton} & & \textbf{Other} & & \textbf{Total} \\
\hline
ARG & & 1,630  & & 1,808  & & 1,319  & & 0      & 1,253  & & 1,443  & & 2,425     & & \textbf{9,878}   \\
AUS & & 0	   & & 0      & & 0      & & 0      & 5,098  & & 4,825  & & 8,938$^*$ & & \textbf{18,861}  \\
BEL & & 1,985  & & 0      & & 0      & & 1,458  & 0      & & 0      & & 2,041     & & \textbf{5,484}   \\
CHN & & 1,344  & & 1,200  & & 1,156  & & 0      & 0      & & 0      & & 1,502$^*$ & & \textbf{5,202}   \\
FRA & & 24,902 & & 21,002 & & 2,294  & & 27,686 & 0      & & 0      & & 32,715    & & \textbf{108,599} \\
GBR & & 4,624  & & 0      & & 0      & & 3,927  & 0      & & 0      & & 4,712     & & \textbf{13,263}  \\
NLD & & 2,388  & & 0      & & 0      & & 2,135  & 0      & & 0      & & 2,795     & & \textbf{7,318}   \\
USA & & 22,224 & & 17,093 & & 17,613 & & 23,226 & 9,897  & & 14,799 & & 28,427    & & \textbf{133,279} \\
\hline
\textbf{Total} & & \textbf{59,097} & & \textbf{41,103} & & \textbf{22,382} & & \textbf{58,432} & \textbf{16,248} & & \textbf{21,067} & & \textbf{83,555} & & \textbf{301,884} \\
\hline
\end{tabular}}
\label{table:cropglobe}
\end{table*}

\subsection{Sample Collection}
\label{sec:samplecollection}
To improve label reliability, we apply a 30-meter inward buffer (based on the CDL resolution) to all reference crop maps after importing them into GEE, which helps remove small noisy patches. As shown in Fig.~\ref{figure:databuffer}(a), erroneous labels within sample areas are removed after buffer filtering. We then match the buffered crop maps with S2 data in both space and time. The intersection of valid crop labels and available cloud-free pixels defines the final sampling region, ensuring that each sampled pixel has both a class label and satellite observations. To reduce sample correlation, the study region is divided into $0.1^\circ \times 0.1^\circ$ cells in geographic coordinates (WGS84), and random samples are drawn within each cell at a spatial resolution of 30 meters, i.e., with a minimum distance of 30 meters between any two sampled points, as shown in Fig.~\ref{figure:databuffer}(b). In the USA, one sample is drawn per cell, whereas in other countries with smaller mapped crop areas, five samples are drawn per cell to balance sample coverage across countries. Taking FRA as an example, a $0.1^\circ \times 0.1^\circ$ cell is approximately 7.7 km $\times$ 11.1 km, and five uniformly random samples with a minimum spacing of 30 meters remain spatially distinct.

Each sample in the constructed CropGlobe dataset includes its geolocation, a full-year time series of acquisition dates, and the corresponding raw reflectance values for the ten spectral bands. The class distribution of the dataset is shown in Table~\ref{table:cropglobe}.

\begin{figure*}[htb!]
\centering
\includegraphics[width=1\textwidth]
{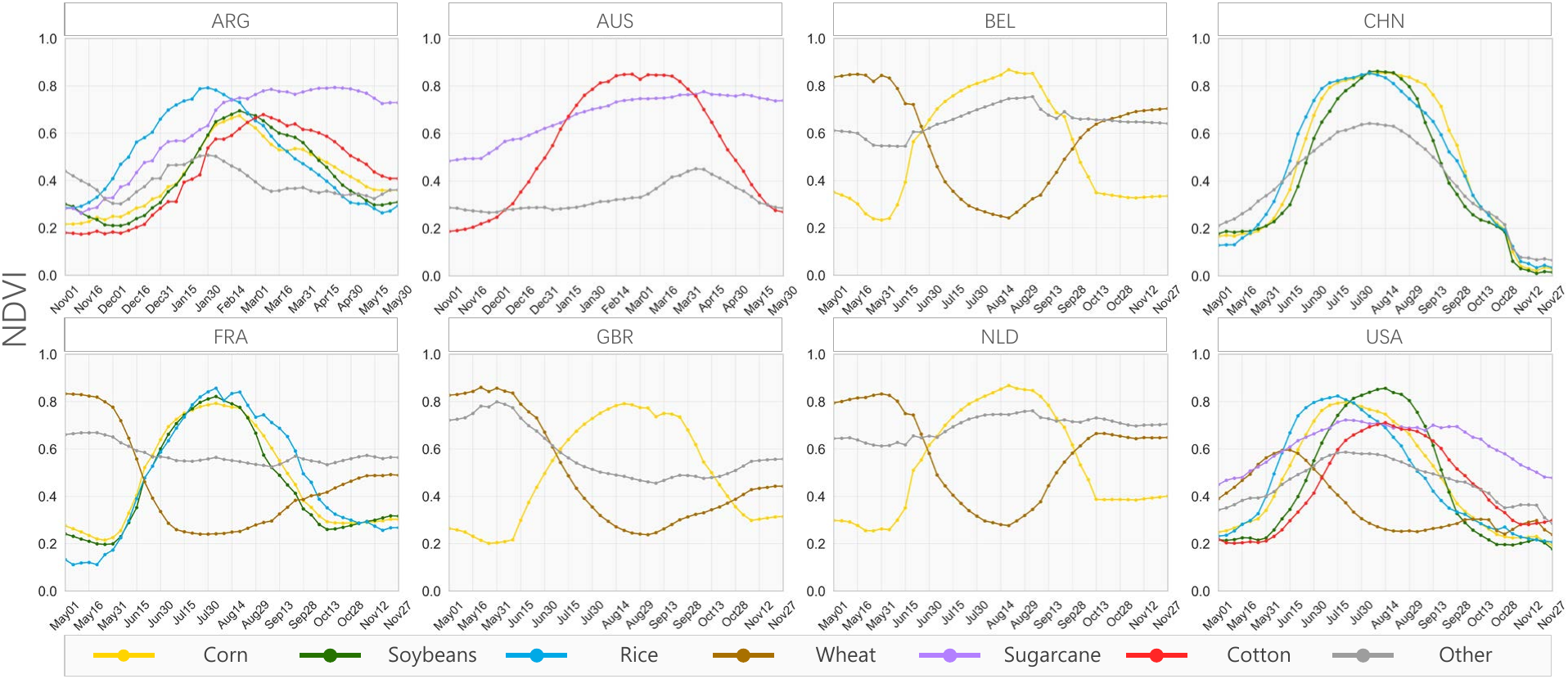}
\caption{Average NDVI curves per country in the CropGlobe dataset. The temporal span covers May--November 2023 for Northern Hemisphere countries and November 2022--May 2023 for Southern Hemisphere countries, with 5-day intervals. The distinct temporal patterns across regions demonstrate the diversity of phenological dynamics in the dataset.}
\label{figure:datandvi}
\end{figure*}

\subsection{Visualization of Crop Time Series by Country}
\label{sec:datavisualization}
To illustrate the diversity and complexity of CropGlobe, we visualize the average Normalized Difference Vegetation Index (NDVI) \cite{NDVI} curves for each country in Fig.~\ref{figure:datandvi}. NDVI is calculated as (NIR - Red)/(NIR + Red) using the near-infrared (NIR; band B8) and red (band B4) bands from S2. We use a 5-day temporal window and focus on May--November 2023 for Northern Hemisphere countries and November 2022--May 2023 for Southern Hemisphere countries, as this period is the most representative for distinguishing crop types according to the sensitivity analysis in Section \ref{sec:resultanalysis}.

The curves show clear cross-country differences in both shape and magnitude. Corn peaks at different times and with different widths across countries, while wheat shows clear temporal misalignment due to differences in planting and harvesting schedules. Peak NDVI values in FRA, GBR, and USA reach around 0.9, whereas lower values are observed in AUS and ARG, possibly reflecting less vigorous canopy growth. Even within the same crop class, curve smoothness, amplitude, and baseline noise vary substantially, highlighting the challenge of feature generalization.

\section{Methodology}
Next we describe feature construction (Section \ref{sec:construction}), the CropNet architecture (Section \ref{sec:cropnet}), data augmentation (Section \ref{sec:augmentation}), and our experimental setup (Section \ref{sec:setup}).

\subsection{Feature Construction}
\label{sec:construction}
To systematically assess the geographic transferability of different remote sensing representations, we construct multiple types of features from S2 observations and geospatial foundation models. Specifically, we consider harmonic features, temporal median features, and GFM embeddings (Presto and AlphaEarth).

\textbf{Harmonic Features:} Harmonic coefficients are derived from four spectral bands, namely Narrow NIR (B8A), shortwave infrared 1 (SWIR 1; B11), SWIR 2 (B12), and NIR (B8). In addition, the Green Chlorophyll Vegetation Index (GCVI = NIR/Green - 1) is included \cite{GCVI}. Frequency-domain features are extracted by applying third-order harmonic regression independently to each spectral variable, resulting in a total of 35 features \cite{GSV1}. This feature design follows prior work \cite{challenge1}, which shows that this combination achieves nearly the same crop-type classification performance as using all spectral bands and a broader set of vegetation indices.

\textbf{Temporal Median Features:} To construct consistent median features from raw S2 observations, we adopt a window-based median compositing strategy. For each pixel, we define a time span of interest bounded by a start day-of-year (DOY) $T_s$ and an end DOY $T_e$, which typically covers the crop growing season. Although observations are collected throughout the year, not all periods contribute equally to crop type classification. In our primary specification for the Northern Hemistphere, we restrict inputs to May--November, which allows the model to focus on the most relevant phenological phases.

The selected time span is then divided into equal-length temporal windows of size $d$, yielding $t = \lceil (T_e - T_s) / d \rceil$ discrete bins, where $\lceil \cdot \rceil$ denotes the ceiling operator. We choose a 5-day compositing interval ($d=5$), which preserves fine-grained phenological dynamics while reducing noise from cloud contamination. For each bin, if valid (cloud-free) observations are available, we compute the median reflectance across all dates within that bin. If no valid observations are available because of cloud contamination, the missing value is filled by linear interpolation along the temporal axis \cite{median2}. If the time series contains too many gaps to support reliable interpolation, the sample is discarded. The final representation is reshaped into a 2D matrix of shape $10 \times t$, where rows represent spectral bands and columns correspond to temporal positions.

\textbf{GFM Embeddings:} We use two representative geospatial foundation models, Presto and AlphaEarth.

Presto \cite{presto} is a pretrained transformer for pixel-level remote sensing time series representation learning. We use the publicly released pretrained Presto model and provide S2 monthly median time series as input, matching the months to our CropNet input for fairness (e.g., May--November for the Northern Hemisphere); more details can be found in Appendix \ref{sec:appendix_presto}. Through self-supervised pretraining, Presto learns to reconstruct masked observations across both temporal positions and sensor modalities, thereby learning unified spectral--temporal representations from large-scale Earth observation data. The extracted Presto embedding is a 128-dimensional feature vector for each pixel.

\begin{sloppypar}
AlphaEarth \cite{Alphaearth} provides globally consistent embeddings of the Earth's surface derived from large-scale multi-source Earth observation data. We directly use the released AlphaEarth embeddings corresponding to each sample location for the year 2023 from the GEE collection \texttt{GOOGLE/SATELLITE\_EMBEDDING/V1/ANNUAL}. Each pixel is represented by a 64-dimensional embedding vector, which serves as a general-purpose learned feature for crop type classification.
\end{sloppypar}

\subsection{CropNet: A Spectral--Temporal Convolutional Architecture}
\label{sec:cropnet}
\begin{figure*}[t]
\centering
\includegraphics[width=0.65\textwidth]
{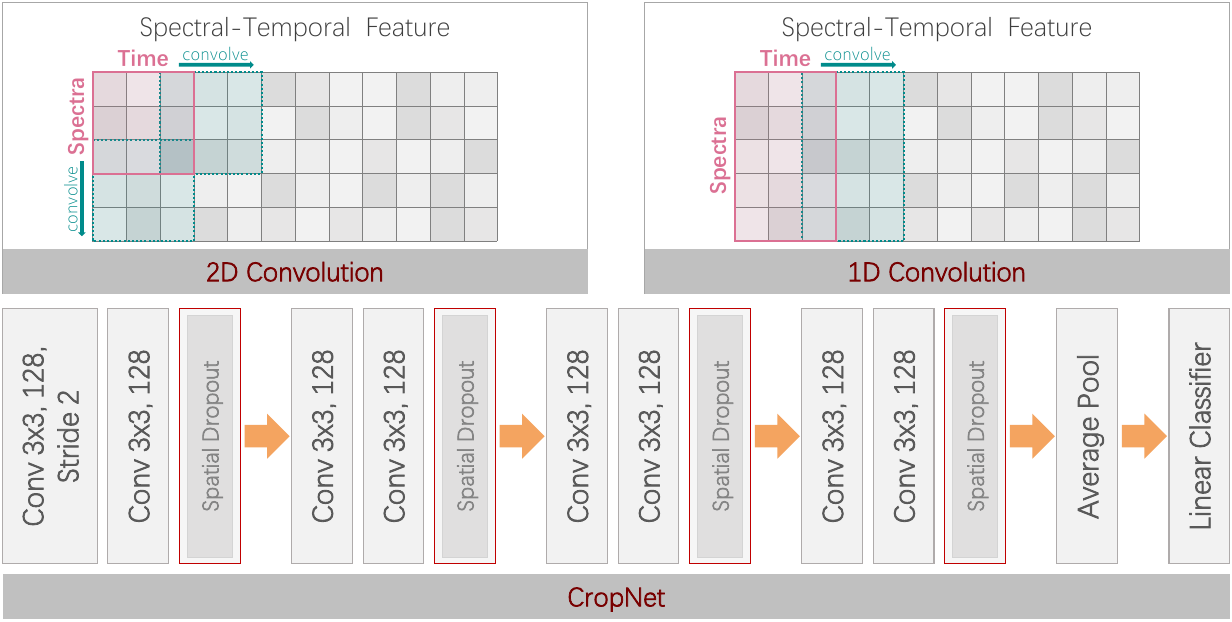}
\caption{Illustration of 2D versus 1D convolution for spectral--temporal features and architecture of CropNet. The network applies 2D convolution jointly along the temporal and spectral dimensions through four convolutional blocks, with spatial dropout used for regularization. Downsampling is performed in the first block. The final crop type prediction is produced by adaptive global average pooling followed by a linear classifier.}
\label{figure:methodcropnet}
\end{figure*}

1D CNNs have been widely used for multispectral time-series classification because they are highly lightweight, computationally efficient, and inherently invariant to shifts along the temporal dimension \cite{tempcnn,1DCNN1,1DCNN2}. However, by treating the multispectral observations at each time step as an indivisible whole and exploring patterns only along the temporal dimension, as shown in Fig.~\ref{figure:methodcropnet}, 1D CNNs may fail to fully exploit the local joint variation of temporal and spectral information throughout the growing season.

To overcome this limitation, we propose CropNet, with two simple yet effective modifications. First, CropNet performs convolution jointly along the temporal and spectral dimensions (Fig.~\ref{figure:methodcropnet}) enabling the network to capture not only local temporal patterns but also local spectral--temporal patterns defined by inter-band interactions. Second, CropNet adopts spatial dropout \cite{SpatialDropout} instead of standard dropout. By suppressing entire feature channels during training, spatial dropout prevents the model from over-relying on a small subset of spectral--temporal responses and encourages predictive evidence to be distributed across multiple channels.
 
Specifically, CropNet consists of four convolutional blocks. Each block contains two 2D convolutional layers with $3 \times 3$ kernels and 128 output channels, with each convolution followed by batch normalization and a Rectified Linear Unit (ReLU) activation. A spatial dropout layer is applied at the end of each block for regularization. Downsampling is performed in the first block through a stride-2 convolution, allowing the network to capture features at different effective scales while remaining computationally efficient.  After the final convolutional stage, adaptive global average pooling reduces the feature map to a $1 \times 1$ representation, which is then fed into a linear classifier for crop type prediction. Overall, these design choices enable CropNet to learn more robust and transferable spectral--temporal representations for cross-region crop type classification.

\subsection{Augmentation of Spectral--Temporal Sequences}
\label{sec:augmentation}
\begin{figure*}[t]
\centering
\includegraphics[width=0.9\textwidth]
{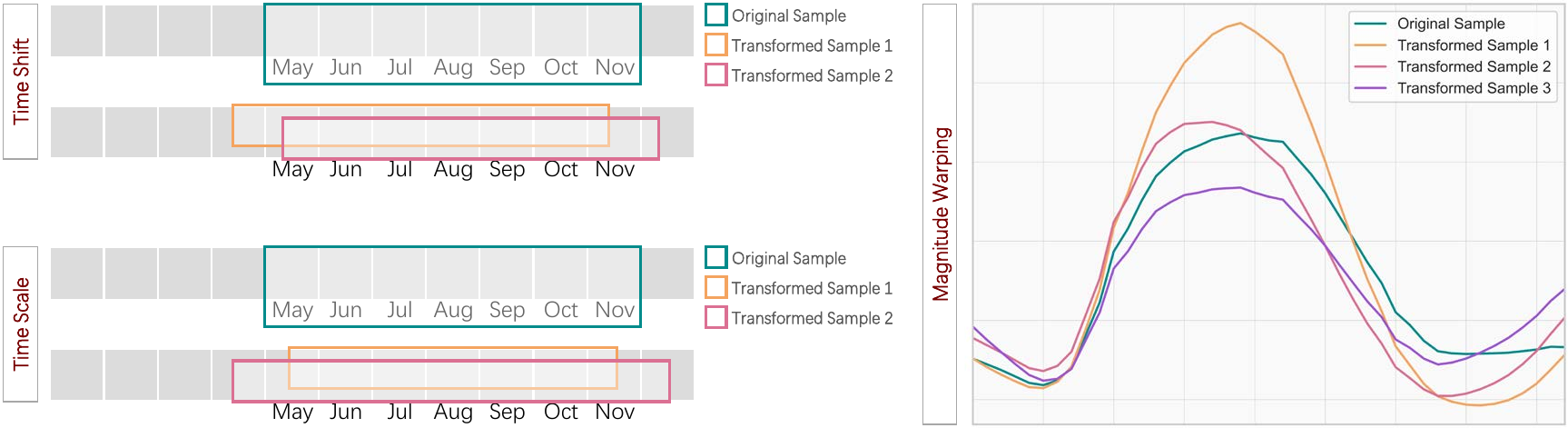}
\caption{Illustration of three transformations for temporal data augmentation: time shift (randomly shifting a fixed temporal window forward or backward), time scale (compressing or stretching the time axis), and magnitude warping (smoothly distorting the amplitude over time, with the same warping applied across spectral bands to preserve their physical relationships).}
\label{figure:methodaugmentation}
\end{figure*}

A major challenge in cross-region crop classification is that the training data often fail to cover the phenological and spectral variation encountered in unseen regions. To mitigate this challenge, we propose a data augmentation strategy tailored to temporal median features. As illustrated in Fig.~\ref{figure:methodaugmentation}, the strategy introduces three transformations: time shift, time scale, and magnitude warping. These transformations are designed to simulate natural variation in crop phenology, observation timing, and reflectance magnitude that commonly occurs across regions and seasons. By exposing the model to such plausible variation during training, the proposed augmentation improves robustness and generalization in cross-region classification.

\textbf{Time Shift} simulates phenological shifts by randomly moving the entire time span forward or backward within a predefined range. Given the original DOY range $[T_s, T_e]$, a random integer offset $\Delta_{\text{shift}}$ is sampled and applied to both the start and end DOYs, resulting in a shifted range $[T_s + \Delta_{\text{shift}}, T_e + \Delta_{\text{shift}}]$. Median values are then recalculated over the shifted time span using the same temporal window size $d$ as in the original setting, thereby preserving feature length consistency. This augmentation accounts for differences in planting dates or climatic conditions across regions.

\textbf{Time Scale} adjusts the temporal extent of the sampling period to capture variability in the duration of crop development. Specifically, the start and end DOYs of the original time span $[T_s, T_e]$ are independently perturbed by random integer offsets $\Delta_{scale}^s$ and $\Delta_{scale}^e$, yielding a scaled range $[T_s', T_e']$, where $T_s' = T_s + \Delta_{\text{scale}}^s$ and $T_e' = T_e + \Delta_{\text{scale}}^e$. Depending on the sampled offsets, the resulting time span may be either expanded or contracted. To maintain a fixed feature dimension, we compute a new temporal window size $d' = \lceil (T_e' - T_s') / t \rceil$, thereby keeping the total number of bins $t$ unchanged. This augmentation mimics differences in growing season length arising from crop varieties, management practices, or climatic conditions.

\textbf{Magnitude Warping} perturbs the temporal curve of each spectral band through a smooth multiplicative modulation, with the aim of mimicking realistic deviations such as sensor noise, sub-pixel heterogeneity, or biomass variability (e.g., the same crop may exhibit different biomass conditions across fields) \cite{MagnitudeWarping1}.
For a given sample and band $b$, let the temporal series be denoted by $\mathbf{x}_b=(x_{b,1},\ldots,x_{b,t})$. Each time point is scaled by a weight $w_i$, such that $x'_{b,i}=w_i\,x_{b,i}$, where the weights are generated from a cubic spline $S(\cdot)$ fitted to $K$ anchor points $(\kappa_j,a_j)$. The anchor abscissae are $\kappa_j = 1 + \frac{j-1}{K-1}(t-1)$ and the ordinates are $a_j \overset{\mathrm{i.i.d.}}{\sim} \mathcal{N}(1,\sigma^2)$ for $j=1,\ldots,K$; the weights are obtained by evaluating the spline at the bin indices $w_i = S(i)$ for $i=1,\ldots,t$ \cite{MagnitudeWarping3, MagnitudeWarping2}.
The same warping transformation is applied to all spectral bands within each sample, that is, the same spline $S(\cdot)$ is shared by every band. This design perturbs the temporal magnitude consistently across bands without changing their relative relationships, thereby maintaining the physical interpretability.

The proposed transformations encourage the model to capture more broadly transferable features, thereby improving robustness to spectral and temporal variability and enabling more stable performance across regions.

\subsection{Experimental Setup}
\label{sec:setup}
We organize our experiments under three progressively more challenging geographic transfer scenarios: cross-country (FRA $\rightarrow$ BEL, FRA $\rightarrow$ NLD, FRA $\rightarrow$ GBR), cross-continent (USA $\rightarrow$ FRA, FRA $\rightarrow$ CHN, FRA $\rightarrow$ USA), and cross-hemisphere (USA $\rightarrow$ AUS, USA $\rightarrow$ ARG, FRA $\rightarrow$ ARG). Here, ``$\rightarrow$'' indicates that the model is trained on the source region and directly applied to the target region without using target-region labels during training. Under these settings, we conduct a systematic comparison of different remote sensing representations to assess their geographic invariance for crop type classification, including harmonic features, median features, and GFM embeddings from Presto and AlphaEarth.

\textbf{Model Comparison:} For harmonic features and GFM embeddings, we evaluate RF with 500 trees and MLP with eight hidden layers, each containing 128 hidden units. For 5-day temporal median features, we evaluate the proposed CropNet as well as a 1D variant obtained by replacing the 2D convolutions in CropNet with 1D convolutions while keeping the remaining settings unchanged. In addition, we compare CropNet with TempCNNs \cite{tempcnn}, a widely used 1D CNN architecture for crop type classification, and several widely used lightweight computer vision baselines, including ResNet50 \cite{ResNet}, EfficientNetV2-S \cite{EfficientNet}, and ConvNeXt-Tiny \cite{ConvNeXt}. To accommodate low-dimensional input features, we reduce the degree of spatial downsampling in these networks to preserve sufficient resolution while keeping their overall architectures unchanged. All neural network models are trained for 50 epochs using the Adam optimizer with a learning rate of 0.0001 and a batch size of 256.

\textbf{Evaluation Metrics:} We report Overall Accuracy (OA) and mean F1 score (mF1) to evaluate classification performance. The mF1 score is computed as the macro-averaged F1 across all categories, reflecting the model's ability to balance class-wise overestimation and underestimation. For each experiment, we repeat training ten times with different random seeds and report the mean and standard deviation.

\textbf{Category Alignment:} Because crop categories differ across countries, we align labels to ensure consistent evaluation. When AUS or CHN (which have three-class crop type maps) are used as target regions, predictions for classes not explicitly defined in their available label set are reassigned to ``Other''. In the FRA $\rightarrow$ USA setting, sugarcane and cotton are excluded from evaluation because they are absent in the source region. No additional alignment is required for other transfer scenarios.

\textbf{Sensitivity Analysis:} Lastly, we conduct sensitivity analysis on the temporal window and time span, as well as ablation experiments on individual data augmentation components, to further examine the impact of these design choices.

\section{Results}
Our results are presented as follows: comparison of different features (Section \ref{sec:results2}); intepretation of CropNet features (Section \ref{sec:results_cam}); comparison of CNN models (Section \ref{sec:resultcnn}); sensitivity analysis and ablation study (Section \ref{sec:resultablation} and Section \ref{sec:resultanalysis}).

\begin{table*}[htb!]
\caption{Crop type classification performance on CropGlobe. Each experiment is repeated ten times, and the mean and standard deviation are reported. ``AUGM'' denotes the use of data augmentation. Values in parentheses indicate performance gain from augmentation. We report only the best-performing classifier for each representation; full RF/MLP results are provided in Appendix \ref{sec:appendix_classifier}.}
\vspace{3mm}
\resizebox{\textwidth}{52mm}{
\begin{tabular}{llrrcrrcrr}
\hline
\multirow{10}{*}{\begin{tabular}[c]{@{}l@{}}Cross-\\ Country\end{tabular}}
& \multirow{2}{*}{Method} & \multicolumn{2}{c}{FRA $\rightarrow$ BEL} && \multicolumn{2}{c}{FRA $\rightarrow$ NLD} && \multicolumn{2}{c}{FRA $\rightarrow$ GBR} \\
\cline{3-4}\cline{6-7}\cline{9-10}
& & OA (\%) & mF1 (\%) & & OA (\%) & mF1 (\%) & & OA (\%) & mF1 (\%) \\
\cline{2-10}
& Harmonic\_MLP	  & 80.86 $\pm$ 0.52 & 84.77 $\pm$ 0.32 && 78.40 $\pm$ 0.81 & 82.47 $\pm$ 0.55 && 62.08 $\pm$ 1.52 & 66.05 $\pm$ 1.36 \\
& Median\_MLP	  & 96.20 $\pm$ 0.71 & 96.79 $\pm$ 0.50 && 93.20 $\pm$ 0.91 & 93.72 $\pm$ 0.82 && 91.84 $\pm$ 0.78 & 92.07 $\pm$ 0.73 \\
& Presto\_RF	  & 96.35 $\pm$ 0.06 & 96.44 $\pm$ 0.06 && 91.09 $\pm$ 0.12 & 91.00 $\pm$ 0.13 && 85.62 $\pm$ 0.42 & 85.89 $\pm$ 0.40 \\
& AlphaEarth\_MLP & 96.99 $\pm$ 0.22 & 97.63 $\pm$ 0.19 && 92.72 $\pm$ 0.65 & 92.81 $\pm$ 0.69 && 93.73 $\pm$ 0.56 & 93.68 $\pm$ 0.56 \\
& CropNet\_1D	  & 96.13 $\pm$ 0.60 & 96.69 $\pm$ 0.70 && 93.73 $\pm$ 0.60 & 93.94 $\pm$ 0.57 && 92.53 $\pm$ 0.77 & 92.85 $\pm$ 0.47 \\
& CropNet	      & \uline{97.74 $\pm$ 0.55} & \uline{97.99 $\pm$ 0.48} && \uline{95.85 $\pm$ 0.43} & \uline{95.98 $\pm$ 0.41} && \uline{95.01 $\pm$ 1.19} & \uline{95.14 $\pm$ 1.19} \\
& CropNet\_AUGM	  & \textbf{98.24 $\pm$ 0.18} & \textbf{98.39 $\pm$ 0.15} && \textbf{96.15 $\pm$ 0.24} & \textbf{96.36 $\pm$ 0.25} && \textbf{95.75 $\pm$ 0.43} & \textbf{95.91 $\pm$ 0.46} \\
&                 & (+ 0.50) & (+ 0.40) & & (+ 0.30) & (+ 0.38) & & (+ 0.74) & (+ 0.77) \\
\hline

\multirow{10}{*}{\begin{tabular}[c]{@{}l@{}}Cross-\\ Continent\end{tabular}}
& \multirow{2}{*}{Method} & \multicolumn{2}{c}{USA $\rightarrow$ FRA} && \multicolumn{2}{c}{FRA $\rightarrow$ CHN} && \multicolumn{2}{c}{FRA $\rightarrow$ USA} \\
\cline{3-4}\cline{6-7}\cline{9-10}
& & OA (\%) & mF1 (\%) & & OA (\%) & mF1 (\%) & & OA (\%) & mF1 (\%) \\
\cline{2-10}
& Harmonic\_MLP	  & 52.20 $\pm$ 0.74 & 54.71 $\pm$ 0.78 && 74.60 $\pm$ 0.89 & 75.07 $\pm$ 0.92 && 53.15 $\pm$ 0.66 & 50.22 $\pm$ 1.05 \\
& Median\_MLP	  & 81.42 $\pm$ 1.86 & 82.41 $\pm$ 1.93 && 44.64 $\pm$ 5.30 & 38.04 $\pm$ 5.61 && 56.43 $\pm$ 1.30 & 51.24 $\pm$ 1.75 \\
& Presto\_RF	  & 62.24 $\pm$ 0.53 & 53.56 $\pm$ 0.88 && 51.94 $\pm$ 0.25 & 34.84 $\pm$ 0.51 && 43.35 $\pm$ 0.32 & 27.11 $\pm$ 0.50 \\
& AlphaEarth\_MLP & 61.74 $\pm$ 2.69 & 61.20 $\pm$ 3.79 && 39.64 $\pm$ 6.54 & 29.46 $\pm$ 9.43 && \uline{79.53 $\pm$ 2.53} & \uline{79.24 $\pm$ 3.69} \\
& CropNet\_1D	  & 82.49 $\pm$ 1.44 & 83.00 $\pm$ 1.17 && 76.94 $\pm$ 0.64 & 76.85 $\pm$ 0.68 && 68.84 $\pm$ 1.77 & 68.84 $\pm$ 1.93 \\
& CropNet	      & \uline{87.00 $\pm$ 0.75} & \uline{89.44 $\pm$ 0.67} && \uline{79.55 $\pm$ 3.20} & \uline{79.31 $\pm$ 3.44} && 73.34 $\pm$ 1.95 & 72.60 $\pm$ 2.21 \\
& CropNet\_AUGM	  & \textbf{87.76 $\pm$ 1.37} & \textbf{90.00 $\pm$ 1.24} && \textbf{82.52 $\pm$ 1.28} & \textbf{82.73 $\pm$ 1.27} && \textbf{80.37 $\pm$ 1.49} & \textbf{80.68 $\pm$ 1.51} \\
&                 & (+ 0.76) & (+ 0.56) & & (+ 2.97) & (+ 3.42) & & (+ 7.03) & (+ 8.08) \\
\hline

\multirow{10}{*}{\begin{tabular}[c]{@{}l@{}}Cross-\\ Hemisphere\end{tabular}}
& \multirow{2}{*}{Method} & \multicolumn{2}{c}{USA $\rightarrow$ AUS} && \multicolumn{2}{c}{USA $\rightarrow$ ARG} && \multicolumn{2}{c}{FRA $\rightarrow$ ARG} \\
\cline{3-4}\cline{6-7}\cline{9-10}
& & OA (\%) & mF1 (\%) & & OA (\%) & mF1 (\%) & & OA (\%) & mF1 (\%) \\
\cline{2-10}
& Harmonic\_MLP	  & 64.52 $\pm$ 1.12  & 64.78 $\pm$ 1.01 && 54.48 $\pm$ 0.70 & 52.92 $\pm$ 0.62 && 45.66 $\pm$ 0.80 & 44.41 $\pm$ 1.00 \\
& Median\_MLP	  & 77.01 $\pm$ 1.93  & 77.58 $\pm$ 1.72 && 67.52 $\pm$ 1.31 & 67.94 $\pm$ 1.43 && 53.24 $\pm$ 2.58 & 49.02 $\pm$ 2.63 \\
& Presto\_RF	  & 45.13 $\pm$ 1.11  & 30.19 $\pm$ 1.19 && 33.36 $\pm$ 3.15 & 21.20 $\pm$ 4.13 && 43.40 $\pm$ 1.63 & 26.98 $\pm$ 0.69 \\
& AlphaEarth\_MLP & 46.31 $\pm$ 10.16 & 30.90 $\pm$ 8.86 && 29.25 $\pm$ 6.46 & 18.29 $\pm$ 5.98 && 28.07 $\pm$ 8.44 & 13.62 $\pm$ 4.36 \\
& CropNet\_1D	  & 82.90 $\pm$ 1.21  & 82.80 $\pm$ 1.15 && 72.06 $\pm$ 0.63 & 72.14 $\pm$ 0.63 && 65.47 $\pm$ 0.95 & 67.95 $\pm$ 0.88 \\
& CropNet	      & \uline{86.74 $\pm$ 1.90}  & \uline{86.75 $\pm$ 1.87} && \uline{74.15 $\pm$ 0.42} & \uline{75.18 $\pm$ 0.45} && \uline{70.92 $\pm$ 2.15} & \uline{70.80 $\pm$ 2.40} \\
& CropNet\_AUGM	  & \textbf{88.45 $\pm$ 0.96}  & \textbf{88.38 $\pm$ 1.02} && \textbf{74.84 $\pm$ 0.79} & \textbf{75.63 $\pm$ 0.79} && \textbf{77.03 $\pm$ 1.01} & \textbf{77.41 $\pm$ 1.16} \\
&                 & (+ 1.71) & (+ 1.63) & & (+ 0.69) & (+ 0.45) & & (+ 6.11) & (+ 6.61) \\
\hline
\end{tabular}}
\label{table:results2}
\end{table*}

\subsection{Temporal Median Features and CropNet Improve Transfer}
\label{sec:results2}
Table~\ref{table:results2} reports performance under progressively more challenging transfer settings. For harmonic, median, Presto, and AlphaEarth representations, we evaluate both RF and MLP; to keep the table concise, we report only the best-performing classifier for each feature type. The full RF/MLP comparison results are provided in Appendix \ref{sec:appendix_classifier}.

\begin{figure*}[t]
\centering
\includegraphics[width=0.9\textwidth]
{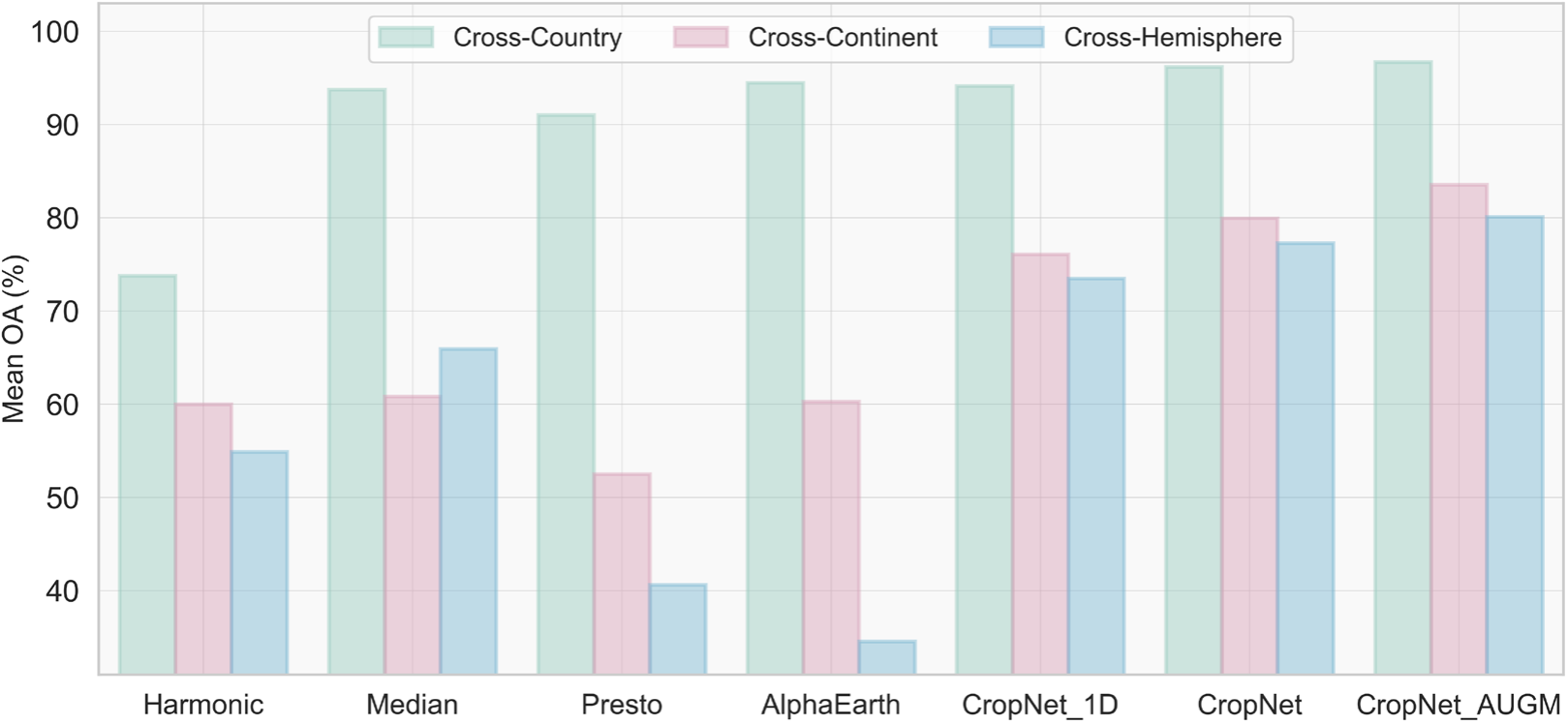}
\caption{Summary of Table~\ref{table:results2} showing the mean OA (\%) of each method under the three transfer settings (cross-country, cross-continent, and cross-hemisphere).}
\label{figure:resultsummary}
\end{figure*}

Overall, performance decreases as the domain shift increases, from cross-country to cross-continent to cross-hemisphere scenarios. Consistent patterns emerge across representations. First, replacing harmonic features with temporal median features yields substantial gains, indicating that preserving fine-grained temporal information facilitates transfer. Second, modeling temporal median features with neural networks further improves performance: MLP-based models outperform RF baselines, and temporal CNNs (CropNet\_1D) improve upon MLPs by introducing temporal convolution. Finally, extending from 1D temporal modeling to joint spectral--temporal convolution (CropNet) yields further gains, achieving the best performance across all transfer settings. For example, in FRA $\rightarrow$ USA, performance increases from 53.2\% (Harmonic\_MLP) to 56.4\% (Median\_MLP) to 68.8\% (CropNet\_1D) and further to 73.3\% (CropNet), demonstrating consistent improvements as stronger inductive biases are introduced. To make the overall trends easier to interpret, we summarize Table~\ref{table:results2} by averaging the OA for each method under the three transfer settings. Fig.~\ref{figure:resultsummary} provides a clearer comparison across methods.

\subsection{CropNet Outperforms Baseline CNN Models}
\label{sec:resultcnn}

\begin{sloppypar}
Table~\ref{table:resultcnn} presents a comparison between CropNet and several widely used CNN baselines, including TempCNN, ResNet50, EfficientNetV2-S, and ConvNeXt-Tiny, for cross-region crop type classification using median features. No data augmentation is applied in these experiments. 
\end{sloppypar}

Compared to TempCNN (0.88M parameters), a strong temporal convolution baseline of similar parameter count, CropNet (1.04M) consistently achieves higher accuracy, indicating that modeling spectral and temporal dimensions jointly provides additional gains beyond temporal modeling alone.

Despite being lightweight, CropNet achieves superior performance compared with much larger vision backbones such as ResNet50 (23.57M), EfficientNetV2-S (20.24M), and ConvNeXt-Tiny (27.85M).
CropNet's advantage is most pronounced under larger domain shifts. For example, in the FRA $\rightarrow$ USA setting, CropNet achieves 73.3\% OA, compared to 60.9\% for ResNet50 and 61.5\% for ConvNeXt-Tiny. Similar gains are observed in cross-hemisphere transfer (e.g., FRA $\rightarrow$ ARG: 70.9\% CropNet vs. 58.4\% for ResNet50).

These results indicate that strong cross-region performance does not require large model capacity, but rather an appropriate inductive bias for modeling spectral--temporal structure.

\begin{figure*}[htb!]
\centering
\includegraphics[width=1\textwidth]
{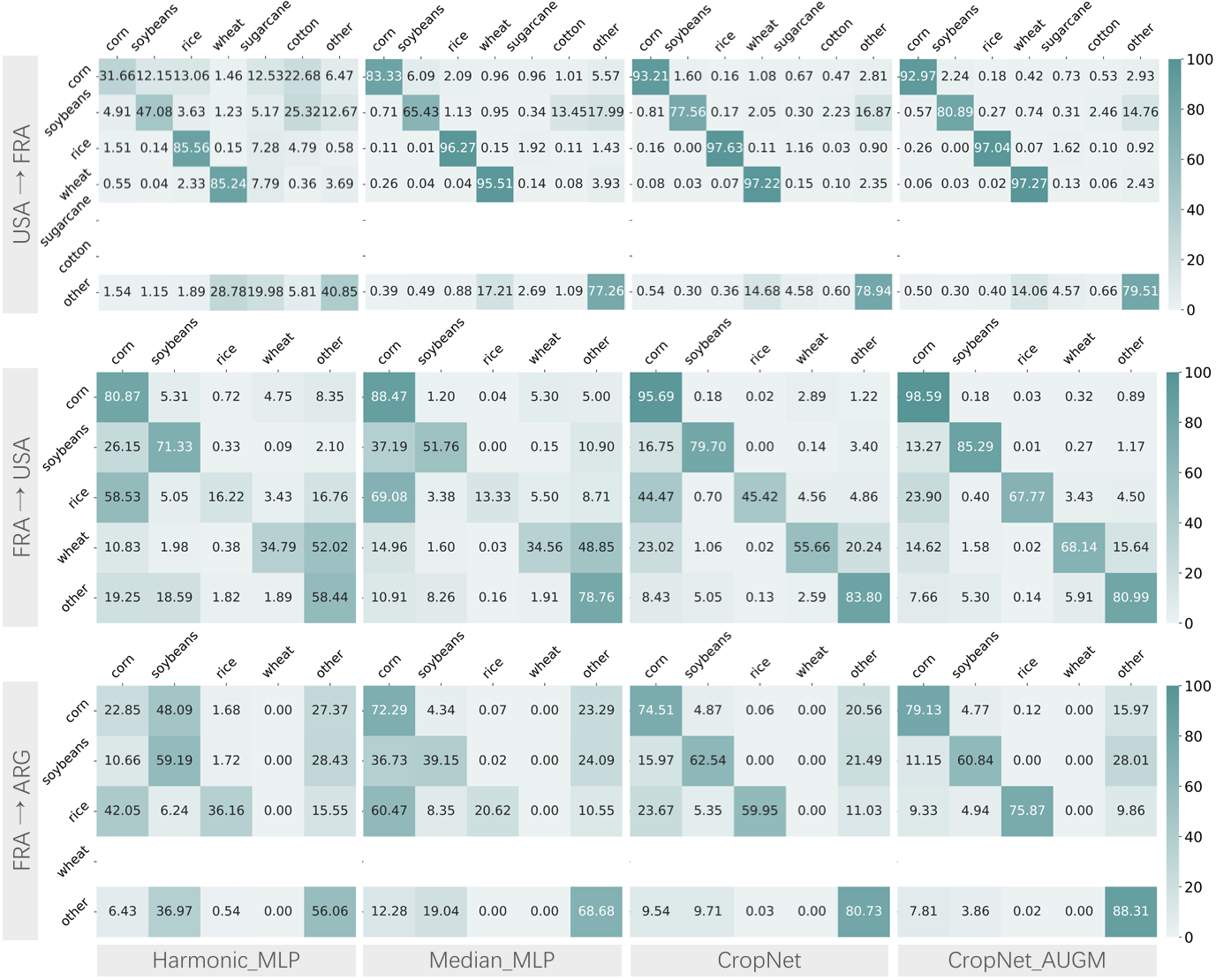}
\caption{Confusion matrices for three representative transfer settings (USA $\rightarrow$ FRA, FRA $\rightarrow$ USA, and FRA $\rightarrow$ ARG) from Table~\ref{table:results2}. Rows correspond to ground-truth crop types and columns to predicted labels.}
\label{figure:resultcm}
\end{figure*}

Fig.~\ref{figure:resultcm} shows confusion matrices for three representative transfer experiments: USA $\rightarrow$ FRA, FRA $\rightarrow$ USA, and FRA $\rightarrow$ ARG, and compares Harmonic\_MLP, Median\_MLP, and CropNet. Overall, temporal median-based models yield clearer diagonals than harmonic features, indicating more reliable cross-region separability. On top of that, CropNet consistently reduces off-diagonal confusion, especially for crop pairs that are easily mixed under larger domain shifts.

The GFM embeddings show a noticeably different behavior. In cross-country transfer, Presto and AlphaEarth can be competitive, indicating that pretrained representations can work well when the source and target domains are relatively close and the shift is mild. However, in cross-continent and especially cross-hemisphere transfer, their performance degrades substantially. 
For example, in the USA $\rightarrow$ AUS setting, Presto achieves 45.1\% and AlphaEarth achieves 46.3\% OA, whereas CropNet reaches 86.7\%, demonstrating a large gap under cross-hemisphere transfer.

\subsection{Data Augmentation Improves Transfer}
\label{sec:resultablation}
Data augmentation improves performance and generally reduces variance, with the largest gains appearing when the source region is limited in diversity.

For example, in the FRA $\rightarrow$ USA setting, augmentation improves OA from 73.3\% to 80.4\%, while in FRA $\rightarrow$ ARG it increases from 70.9\% to 77.0\%.

Although we sample France at higher density, its geographic extent is still relatively limited, which constrains the diversity of crop calendars and background conditions seen during training. 
When the target region exhibits larger phenological and radiometric shifts than seen in the source region, augmentation helps compensate for the lack of diversity by exposing the model to plausible temporal and magnitude variations. In comparison, when training on the USA, the benefit of augmentation is smaller (e.g., USA $\rightarrow$ AUS: 86.7\% to 88.5\%). This is likely because USA already spans a wide range of climates and management regimes, providing a more diverse source distribution.

We conduct an ablation study on the data augmentation components (Table~\ref{table:analysisablation}) to understand their relative contributions. The experiments are conducted under challenging transfer settings where the source domain is relatively limited in diversity (FRA $\rightarrow$ USA and FRA $\rightarrow$ ARG) or performance is otherwise lower (USA $\rightarrow$ AUS).
Time shift provides the largest improvement over the no-augmentation baseline, yielding substantial gains across all settings. Adding time scale and magnitude warping each leads to smaller additional improvements. For example, in FRA $\rightarrow$ USA, performance improves from 73.3\% (no augmentation) to 79.4\% with time shift and to 80.5\% with time shift and time scale, while adding magnitude warping yields comparable performance.

\begin{table*}[t]
\centering
\caption{Comparison of CNN models for crop type classification on CropGlobe. All models operate on median features and no data augmentation is applied.}
\vspace{3mm}
\resizebox{0.9\textwidth}{19mm}{
\begin{tabular}{llcrccrcrcrcr}
\hline
\multicolumn{2}{l}{CNN Model} & & TempCNNs & & & ResNet50 & & EfficientNetV2-S & & ConvNeXt-Tiny & & CropNet (Ours) \\
\cline{1-2}
\multicolumn{2}{l}{Parameters} & & 0.88 M & & & 23.57 M & & 20.24 M & & 27.85 M & & 1.04 M \\
\hline
\multirow{3}{*}{\begin{tabular}[c]{@{}l@{}}Cross-\\ Country\end{tabular}}
& FRA $\rightarrow$ BEL & & 97.70 $\pm$ 0.43  & & & 96.57 $\pm$ 1.09 & & 97.28 $\pm$ 0.51 & & 97.22 $\pm$ 0.59 & & \textbf{97.74 $\pm$ 0.55} \\
& FRA $\rightarrow$ NLD & & 95.52 $\pm$ 0.23  & & & 93.16 $\pm$ 0.67 & & 94.14 $\pm$ 1.09 & & 94.10 $\pm$ 0.54 & & \textbf{95.85 $\pm$ 0.43} \\
& FRA $\rightarrow$ GBR & & 94.21 $\pm$ 0.65  & & & 91.91 $\pm$ 1.44 & & 93.04 $\pm$ 2.12 & & 93.20 $\pm$ 0.68 & & \textbf{95.01 $\pm$ 1.19} \\
\hline

\multirow{3}{*}{\begin{tabular}[c]{@{}l@{}}Cross-\\ Continent\end{tabular}}
& USA $\rightarrow$ FRA & & 85.94 $\pm$ 0.71  & & & 81.46 $\pm$ 1.88 & & 84.09 $\pm$ 1.90 & & 81.36 $\pm$ 1.66 & & \textbf{87.00 $\pm$ 0.75} \\
& FRA $\rightarrow$ CHN & & 66.17 $\pm$ 10.29 & & & 39.46 $\pm$ 3.80 & & 69.37 $\pm$ 6.71 & & \textbf{80.01 $\pm$ 1.88} & & 79.55 $\pm$ 3.20 \\
& FRA $\rightarrow$ USA & & 62.86 $\pm$ 1.01  & & & 60.89 $\pm$ 1.79 & & 64.52 $\pm$ 1.91 & & 61.49 $\pm$ 2.55 & & \textbf{73.34 $\pm$ 1.95} \\
\hline

\multirow{3}{*}{\begin{tabular}[c]{@{}l@{}}Cross-\\ Hemisphere\end{tabular}}
& USA $\rightarrow$ AUS & & 84.04 $\pm$ 0.94  & & & 78.56 $\pm$ 2.74 & & 82.02 $\pm$ 2.13 & & 78.63 $\pm$ 3.59 & & \textbf{86.74 $\pm$ 1.90} \\
& USA $\rightarrow$ ARG & & 70.86 $\pm$ 0.44  & & & 68.07 $\pm$ 1.12 & & 69.39 $\pm$ 0.92 & & 64.21 $\pm$ 1.63 & & \textbf{74.15 $\pm$ 0.42} \\
& FRA $\rightarrow$ ARG & & 60.65 $\pm$ 1.48  & & & 58.42 $\pm$ 2.88 & & 58.93 $\pm$ 3.48 & & 61.27 $\pm$ 2.37 & & \textbf{70.92 $\pm$ 2.15} \\
\hline
\end{tabular}}
\label{table:resultcnn}
\end{table*}

\begin{table*}[t]
\centering
\caption{Ablation study on the components of data augmentation. We evaluate the individual and combined impact of time shift, time scale, and magnitude warping.}
\vspace{3mm}
\resizebox{\textwidth}{11mm}{
\begin{tabular}{lrrcrrcrr}
\hline
\multirow{2}{*}{Components} & \multicolumn{2}{c}{FRA $\rightarrow$ USA} & & \multicolumn{2}{c}{FRA $\rightarrow$ ARG} & & \multicolumn{2}{c}{USA $\rightarrow$ AUS} \\
\cline{2-3}\cline{5-6}\cline{8-9}
& OA (\%) & mF1 (\%) & & OA (\%) & mF1 (\%) & & OA (\%) & mF1 (\%) \\
\hline
No Augmentation	
& 73.34 $\pm$ 1.95 & 72.60 $\pm$ 2.21 & & 70.92 $\pm$ 2.15 & 70.80 $\pm$ 2.40 & & 86.74 $\pm$ 1.90 & 86.75 $\pm$ 1.87 \\
Time Shift	
& 79.38 $\pm$ 1.20 & 79.65 $\pm$ 1.40 & & 75.66 $\pm$ 1.01 & 75.86 $\pm$ 1.24 & & 87.33 $\pm$ 1.70 & 87.20 $\pm$ 1.77 \\
Time Shift + Time Scale	
& \textbf{80.48 $\pm$ 2.34} & 80.66 $\pm$ 2.49 & & 76.01 $\pm$ 0.69 & 76.41 $\pm$ 0.79 & & 87.70 $\pm$ 0.85 & 87.60 $\pm$ 0.90 \\
Time Shift + Time Scale + Magnitude Warping	
& 80.37 $\pm$ 1.49 & \textbf{80.68 $\pm$ 1.51} & & \textbf{77.03 $\pm$ 1.01} & \textbf{77.41 $\pm$ 1.16} & & \textbf{88.45 $\pm$ 0.96} & \textbf{88.38 $\pm$ 1.02} \\
\hline
\end{tabular}}
\label{table:analysisablation}
\end{table*}

\subsection{Interpreting CropNet's Learned Spectral--Temporal Features}
\label{sec:results_cam}
To better understand what drives CropNet's predictions, as well as the reason for its advantage over CropNet\_1D, we use Class Activation Maps (CAMs) \cite{cams} to visualize the input regions that contribute most to classification. Specifically, we compute CAMs for both CropNet and CropNet\_1D by backpropagating gradients from the predicted class to intermediate convolutional layers, producing importance patterns distributed over spectral--temporal features.

\begin{figure*}[t]
\centering
\includegraphics[width=1\textwidth]
{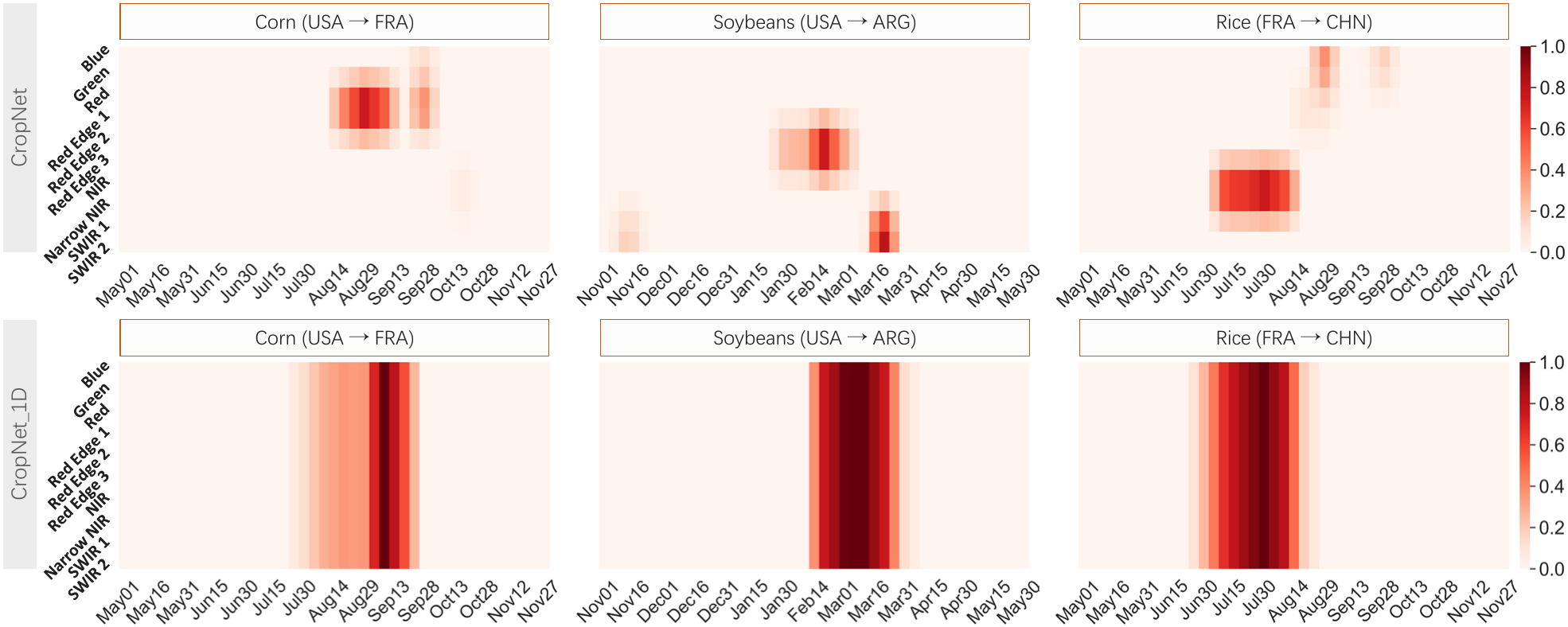}
\caption{Visualization of feature importance maps from CropNet and CropNet\_1D for three crop and transfer settings. The x-axis indicates time, and the y-axis shows S2 spectral bands. For ARG, time has been shifted six months earlier. The highlighted areas indicate the most informative feature regions for classification, illustrating how 2D convolution and 1D convolution capture crop-discriminative cues differently.}
\label{figure:resultfeatureall}
\end{figure*}

Fig.~\ref{figure:resultfeatureall} shows the CAM-based importance maps for three crop types, revealing differences in how the two convolutional architectures organize discriminative information. The high-response regions of CropNet are not distributed only along the temporal axis, but form compact local regions in the spectral--temporal plane. This indicates that 2D convolutions are able to capture joint variation patterns among several key spectral bands within specific temporal windows. At a given growth stage, what is truly discriminative is often not that all bands are important simultaneously, but that a subset of specific bands, together with that time period, forms a stable basis for recognition. The figure also shows that the most informative bands are consistently concentrated in the Red, Red Edge, NIR, and SWIR ranges (Red, Red Edge 1/2/3, Narrow NIR, SWIR 1/2). These bands are known to be sensitive to chlorophyll content, canopy structure, and water content, which are key biophysical factors that differentiate crop growth trajectories. 

In contrast, the high responses of CropNet\_1D mainly appear as several relatively broad temporal intervals and lack clear spectral structure. This suggests that CropNet\_1D primarily learns when informative signals occur, but is less able to express which spectral bands are most important at those times.

\begin{figure*}[t!]
\centering
\includegraphics[width=1\textwidth]
{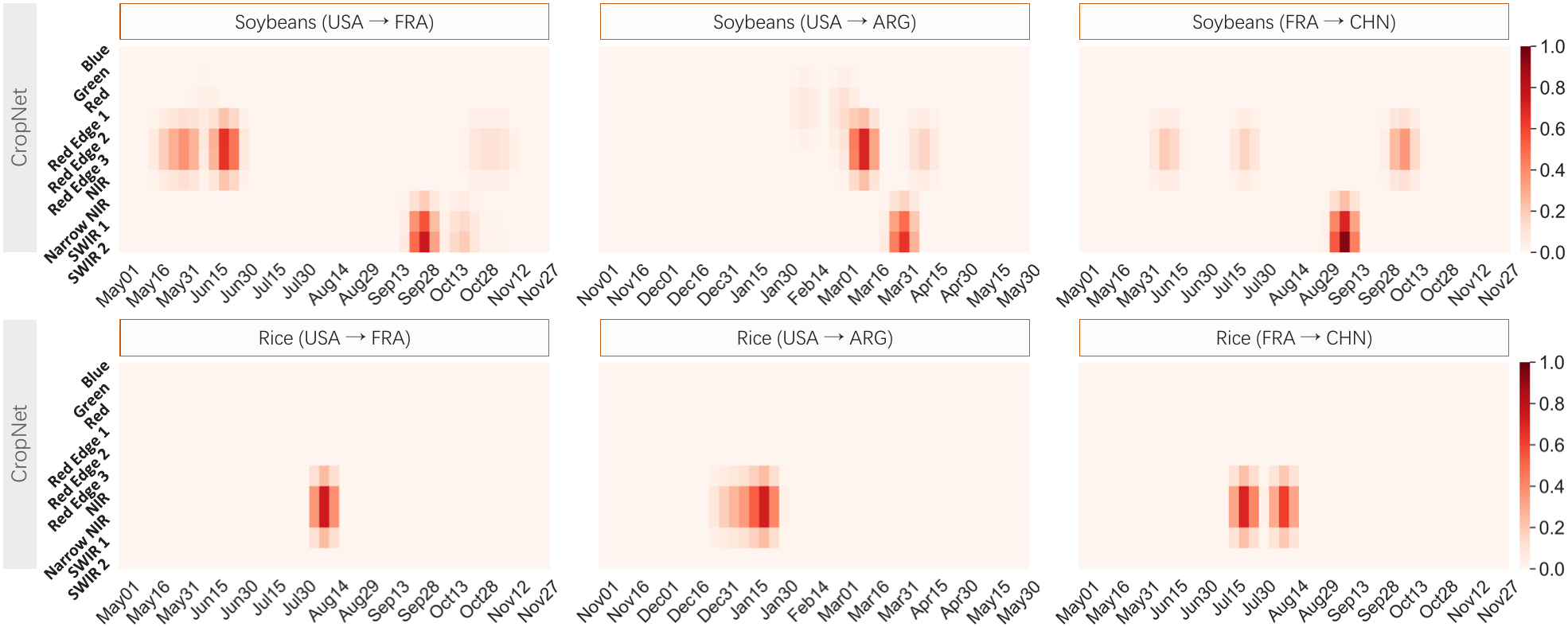}
\caption{Feature importance maps of the same crop (soybeans and rice) in different target regions. The x-axis denotes time, and the y-axis indicates S2 spectral bands. While the time importance is distributed across different bins due to region-specific crop growth phenology, the most informative spectral bands remain consistent across regions.}
\label{figure:resultfeaturesame}
\end{figure*}

To further examine cross-region generalization, we visualize CAMs for the same crop (soybeans and rice) under different target regions (Fig.~\ref{figure:resultfeaturesame}). For a given crop, the most important time periods shift across regions, consistent with variations in planting dates and growth rates, while the key spectral bands remain largely stable. For soybeans, the most informative bands are typically from Red through Red Edge 2 and from Narrow NIR through SWIR 2, capturing changes in photosynthetic activity and moisture stress. For rice, importance concentrates more strongly around Red Edge 3 to SWIR 1, which is responsive to canopy structure and water-related dynamics associated with paddy systems. Together, these results suggest that while phenology is region-dependent, the underlying spectral sensitivities exploited by the model are more consistent, enabling shared cues to transfer across geographic domains.

\subsection{Sensitivity Analysis}
\label{sec:resultanalysis}
We examine how temporal window and time span affect transfer performance. Shorter compositing windows consistently perform better, with 5-day intervals providing the strongest results across all transfer settings, highlighting the importance of preserving fine-grained phenological dynamics. The effect of time span becomes more pronounced under larger domain shifts: performance is relatively stable for cross-country transfer, but more sensitive for cross-continent and cross-hemisphere transfer. Including early-year observations may introduce mismatched vegetation signals, whereas restricting inputs to the main growing season improves robustness. Overall, a May--November time span provides the best trade-off across settings and is therefore adopted in all main experiments. 

We also assess sensitivity to data augmentation parameters. Moderate perturbation ranges consistently yield the best performance. In particular, time shift and time scale are most effective under moderate ranges ($[-30, +30]$), while magnitude warping performs best when introducing mild, smooth reflectance variation ($\sigma=0.1$, $knots=5$). These results suggest that augmentation is most beneficial when it reflects realistic but bounded variation in crop growth dynamics. 

These conclusions are supported by detailed analyses of temporal window, time span, and data augmentation parameters, which are provided in Appendix \ref{sec:appendix_sensitivity}.

\section{Discussion}
\subsection{What Enables Geographic Transfer in Crop Type Mapping?}
Our results reveal that the invariant signatures of major crops reside in the coupled spectral--temporal domain. A key finding is that S2 temporal features (5-day medians) transfer better than commonly assumed. Even with a lightweight 1D temporal CNN, this representation delivers strong baseline accuracy across cross-continent and cross-hemisphere settings. This is likely because temporal convolution is inherently tolerant to shifts along the time axis, allowing the model to recognize phenological patterns even when planting dates vary between regions.

At the same time, CropNet's 2D spectral--temporal convolutions provide an even better inductive bias for geographic transfer. As shown in Table~\ref{table:results2}, CropNet consistently outperforms 1D temporal CNNs in all transfer settings. 
Why do 2D convolutions offer an advantage?
A natural hypothesis is that convolving over the spectral axis could introduce a form of spectral shift invariance, allowing the model to recognize crop signals even when reflectance drifts across adjacent bands due to atmospheric or background effects. However, our CAM analysis (Figure~\ref{figure:resultfeaturesame}) does not provide strong evidence for such cross-band shifts. 

A more plausible explanation is that joint convolution more effectively captures coupled physical relationships across spectral bands over time. A 1D model treats all bands as a whole, and the spectral dimension is fully compressed after convolution. In contrast, a 2D architecture continuously maintains features in a spectral--temporal form. By applying shared kernels across both time and wavelength, CropNet can learn the most critical local spectral--temporal correlations --- such as coordinated changes in the Red Edge bands --- which appear to be more stable across regions.

We also observe that GFM embeddings are not uniformly reliable for crop type transfer. In relatively mild shifts (e.g., cross-country transfer within Western Europe), Presto and AlphaEarth can be competitive. However, their performance degrades substantially in more challenging cross-continent and cross-hemisphere settings. 
This suggests that the issue is not necessarily a lack of discriminative power, but rather the encoding of geographically dependent features that act as effective ``shortcuts'' for in-region classification but become confounding noise in distant regions. 
More broadly, this indicates that self-supervised reconstruction objectives do not inherently enforce invariance to geographic shifts and permit the retention of region-specific signals that hinder transfer.

Beyond representation and model design, temporal data augmentation further improves robustness. Strategies such as time shift, time scale, and magnitude warping yield the largest gains when the source region is limited and the target domain shift is large. This suggests that augmentation can partially compensate for insufficient source diversity by exposing the model to plausible variations in crop growth dynamics.

\subsection{When Does Transfer Work, and When Does it Break Down?}
Our results show that cross-region crop mapping is feasible in many settings without local labels when appropriate spectral--temporal representations are used. These findings suggest that sharing data and models across regions can enable rapid deployment in new areas and reduce annotation costs for maintaining existing maps. In addition, we expect similar benefits to extend across years (e.g., under anomalous seasons). We report USA cross-temporal transfer results in Appendix \ref{sec:appendix_crosstime}, but cross-year transfer is not the focus of this work.

\begin{figure*}[htb!]
\centering
\includegraphics[width=1\textwidth]
{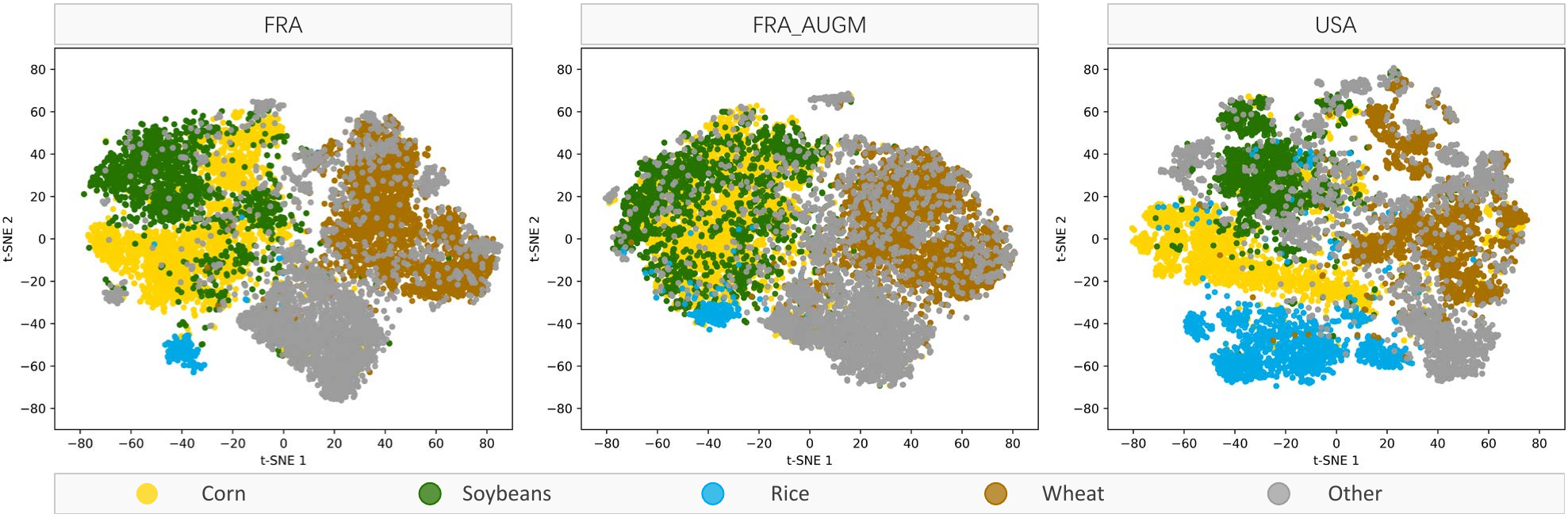}
\caption{t-SNE visualization of S2 median features (May--November; 5-day intervals) for FRA (pre-/post-augmentation) and USA. For comparability, USA is restricted to the same five classes as FRA.}
\label{figure:discussiondistribution}
\end{figure*}

Notably, transfer performance is influenced not only by the representation and model, but also strongly by the diversity of the training data. To better understand this effect, we visualize feature distributions using t-SNE \cite{t-SNE} for FRA (before and after augmentation) and USA (Fig.~\ref{figure:discussiondistribution}). USA samples are more dispersed, with weaker class separation, indicating greater variability in spectral--temporal patterns. In contrast, FRA forms tighter clusters, reflecting lower within-class variability but also narrower coverage of real-world conditions. Augmentation expands the FRA distribution and increases within-class diversity, but only within a limited range.

This reveals a key constraint: transfer performance is limited by the diversity of the source domain. When the source region does not capture the range of spectral--temporal variation present in the target, models fail to generalize, even if the underlying representation is strong.

Geographic coverage remains another limitation of our study. Our evaluation focuses primarily on relatively high-income countries. These settings differ systematically from many low-income regions; most notably, intercropping is uncommon in our study countries but prevalent in Sub-Saharan Africa and parts of South Asia. Yield levels and variability also differ markedly across these contexts (e.g., due to inputs, cultivars, irrigation, and field structure), which can shift spectral--temporal signatures and label noise. These differences limit truly global applicability, and direct transfer to such regions will likely be more challenging.

\subsection{Does Transfer Performance Correlate with Regional Yield Levels?}
To further probe what drives transferability, we examine whether crop productivity influences the invariance of spectral--temporal features across regions. The underlying hypothesis is that transferability may be higher between regions with more similar yields, particularly when both are high yielding, because in such conditions S2 reflectance is less influenced by soil, and management practices across regions tend to be more uniform.

\begin{figure*}[htb!]
\centering
\includegraphics[width=1\textwidth]
{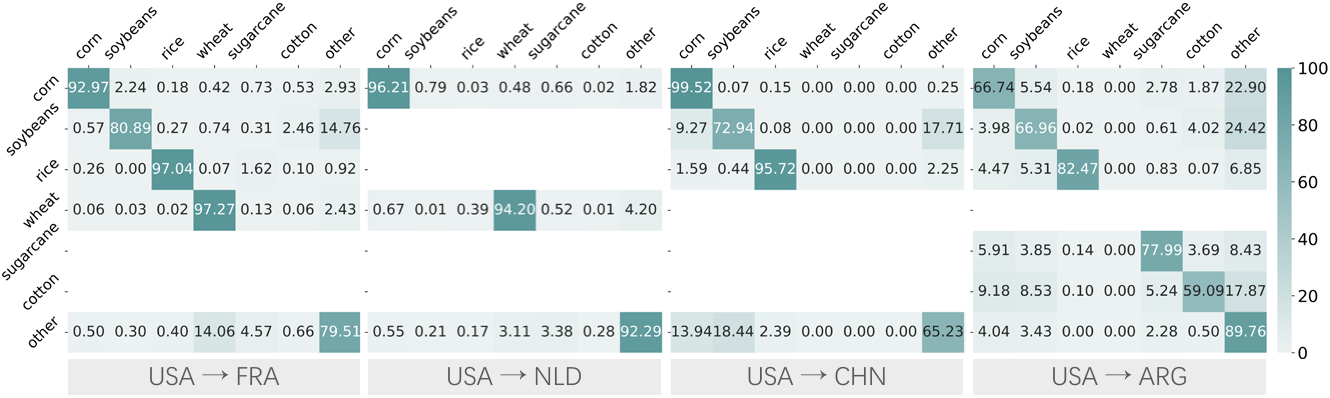}
\caption{Confusion matrices of CropNet with data augmentation for four cross-region transfer settings, used to analyze the potential relationship between classification performance and crop yields.}
\label{figure:discussioncm}
\end{figure*}

We collect 2023 crop yield statistics from the Food and Agriculture Organization of the United Nations (\url{https://www.fao.org/home/en}) and summarize them in Table~\ref{table:discussionyield}. We then compare these with classification results from selected transfer settings, as shown in the confusion matrices in Fig.~\ref{figure:discussioncm}.

For rice, yield differences between USA (8572.8 kg/ha) and CHN (7132.5 kg/ha) or ARG (6723.7 kg/ha) are smaller than those between USA and FRA (5632.8 kg/ha), yet USA $\rightarrow$ FRA shows better classification accuracy than USA $\rightarrow$ CHN or USA $\rightarrow$ ARG.

For wheat, despite USA having much lower yield (3269.4 kg/ha) than FRA (7201.6 kg/ha) and NLD (8467.0 kg/ha), classification accuracy remains high in both transfer directions. This shows that, at least under the roughly twofold yield gap in this example, yield differences do not prevent high performance in cross-continent transfer.

Taken together, within the yield ranges represented in our study (mostly modest, except for approximately twofold differences in wheat), we do not find a consistent relationship between yield levels and cross-region accuracy; other factors, such as training-data quantity and quality, class balance, and the spectral--temporal distinctiveness of the crop types, appear to matter more. At the same time, our analysis cannot rule out yield as a dominant factor when gaps are much larger (e.g., in low-input smallholder systems typical of parts of Sub-Saharan Africa), where lower canopy density and greater soil/background influence may substantially alter S2 signals. Evaluating such settings requires data that span a broader yield spectrum and will be a direction for future work.

\begin{table*}[htb!]
\centering
\caption{Crop yields (kg/ha) for target crop types across countries in 2023.}
\vspace{3mm}
\resizebox{0.5\textwidth}{13mm}{
\begin{tabular}{lrrrrrr}
\hline
\textbf{Yield (kg/ha)} & \textbf{Corn} & \textbf{Soybeans} & \textbf{Rice} & \textbf{Wheat} & \textbf{Sugarcane} & \textbf{Cotton} \\
\hline
ARG & 5109.4  & 1744.5 & 6723.7 & --     & 30309.6 & 1751.6 \\
AUS & --      & --     & --     & --     & 98619.5 & 3787.4 \\
BEL & 9828.9  & --     & --     & 8424.7 & --      & --	    \\
CHN & 6532.0  & 1952.6 & 7132.5 & --     & --      & --	    \\
FRA & 9762.5  & 2458.4 & 5632.8 & 7201.6 & --      & --	    \\
GBR & --      & --     & --     & 8127.9 & --      & --	    \\
NLD & 10265.5 & --     & --     & 8467.0 & --      & --	    \\
USA & 11130.6 & 3398.7 & 8572.8 & 3269.4 & 79309.2 & 2843.0 \\
\hline
\end{tabular}}
\label{table:discussionyield}
\end{table*}

\subsection{How Can Transferability Reach the Regions Most in Need?}
An especially important next step is to transfer crop type mapping tools to regions that are most food insecure and in urgent need of reliable agricultural statistics. For instance, many countries in sub-Saharan Africa remain chronically data-scarce despite being highly vulnerable to food crises \cite{africa1, africa2}. Our results show that cross-hemisphere transfer is feasible: being in the Southern Hemisphere is not a barrier. The key question is when transfer fails. 

The yield range in the CropGlobe dataset does not reach the lower levels common in many African systems: corn yields in some areas are less than 5 t/ha (CropGlobe's lowest is ARG at approximately 5.1 t/ha), and rice yields are often less than 2 t/ha (FRA in our dataset is about 5.6 t/ha). Under such low-yield (low-biomass) conditions, combined with differences in phenology and environment, models trained on our data may not transfer directly to these more challenging regions. 

Even so, our results already show that S2 spectral--temporal features achieve consistently strong cross-country accuracy, and we hypothesize a feasible path forward. By collecting crop type data from neighboring or similar countries to supplement the training data, the model can be exposed to spectral--temporal patterns close to those expected in the target. Approaches such as soft labels \cite{weakly1, weakly2}, label transfer \cite{labeltransfer}, and weak constraints \cite{weakconstraints} can be used to anchor the model to the target region’s distribution and require only a small amount of local ground truth for calibration. This path does not require the same quantity or quality of labels as in our study, but can improve robustness and, while moving toward operational mapping, greatly reduce local labeling needs and costs.

The transfer learning method developed here is able to serve as a first step. On this basis, future work will extend to more challenging settings, such as lower-yield, intercropped, and highly heterogeneous systems. A longer-term aim is to enable operational crop type mapping in low-income, data-scarce countries and to lower labeling and statistical costs. This is a promising route to bring cost-effective crop type mapping to the regions that need it most.

\section{Conclusion}
This study systematically investigates the geographic invariance of various remote sensing representations for cross-region crop type classification. By constructing the CropGlobe dataset, which spans eight countries across five continents, we conduct experiments under three transfer settings: cross-country, cross-continent, and cross-hemisphere. The results show that the 5-day temporal median features derived from S2 exhibit significant advantages in geographic transferability.
To better capture the spectral--temporal structure, we introduce CropNet, a lightweight CNN that outperforms mainstream convolutional models, indicating that an appropriate inductive bias is more important than model scale for cross-region crop classification.
We further show that augmentation of temporal median features improves robustness to geographic variation. By simulating realistic shifts in crop phenology and reflectance magnitude, the proposed augmentation strategy yields substantial gains when the source region is limited in diversity and the target domain shift is large.
Our work provides initial evidence that global crop type mapping can be practical in many settings with fewer local ground samples, and it offers a data resource and a methodological framework for studying feature invariance in crop type identification. Looking ahead, as the dataset expands and methods evolve, this study could help move from region-specific applications toward more broadly generalizable solutions, with the potential to support agricultural monitoring, food-security planning, and policy analysis.

\bibliographystyle{IEEEtran}
\bibliography{reference}

@article{importance1,
  title={Farming the planet: 2. Geographic distribution of crop areas, yields, physiological types, and net primary production in the year 2000},
  author={Monfreda, Chad and Ramankutty, Navin and Foley, Jonathan A},
  journal={Global Biogeochemical Cycles},
  volume={22},
  number={1},
  year={2008},
  publisher={Wiley Online Library}
}

@article{importance2,
  title={Strengthening agricultural decisions in countries at risk of food insecurity: The GEOGLAM Crop Monitor for Early Warning},
  author={Becker-Reshef, Inbal and Justice, Christina and Barker, Brian and Humber, Michael and Rembold, Felix and Bonifacio, Rogerio and Zappacosta, Mario and Budde, Mike and Magadzire, Tamuka and Shitote, Chris and others},
  journal={Remote Sensing of Environment},
  volume={237},
  pages={111553},
  year={2020},
  publisher={Elsevier}
}

@article{importance3,
  title={The influence of climate change on global crop productivity},
  author={Lobell, David B and Gourdji, Sharon M},
  journal={Plant Physiology},
  volume={160},
  number={4},
  pages={1686--1697},
  year={2012},
  publisher={American Society of Plant Biologists}
}

@article{importance4,
  title={Dynamic adaptive policy pathways: A method for crafting robust decisions for a deeply uncertain world},
  author={Haasnoot, Marjolijn and Kwakkel, Jan H and Walker, Warren E and Ter Maat, Judith},
  journal={Global Environmental Change},
  volume={23},
  number={2},
  pages={485--498},
  year={2013},
  publisher={Elsevier}
}

@article{importance5,
  title={Solutions for a cultivated planet},
  author={Foley, Jonathan A and Ramankutty, Navin and Brauman, Kate A and Cassidy, Emily S and Gerber, James S and Johnston, Matt and Mueller, Nathaniel D and O’Connell, Christine and Ray, Deepak K and West, Paul C and others},
  journal={Nature},
  volume={478},
  number={7369},
  pages={337--342},
  year={2011},
  publisher={Nature Publishing Group UK London}
}

@article{challenge1,
  title={Crop type mapping without field-level labels: Random forest transfer and unsupervised clustering techniques},
  author={Wang, Sherrie and Azzari, George and Lobell, David B},
  journal={Remote Sensing of Environment},
  volume={222},
  pages={303--317},
  year={2019},
  publisher={Elsevier}
}

@article{challenge2,
  title={Two shifts for crop mapping: Leveraging aggregate crop statistics to improve satellite-based maps in new regions},
  author={Kluger, Dan M and Wang, Sherrie and Lobell, David B},
  journal={Remote Sensing of Environment},
  volume={262},
  pages={112488},
  year={2021},
  publisher={Elsevier}
}

@article{challenge3,
  title={Automated in-season mapping of winter wheat in China with training data generation and model transfer},
  author={Yang, Gaoxiang and Li, Xingrong and Liu, Pengzhi and Yao, Xia and Zhu, Yan and Cao, Weixing and Cheng, Tao},
  journal={ISPRS Journal of Photogrammetry and Remote Sensing},
  volume={202},
  pages={422--438},
  year={2023},
  publisher={Elsevier}
}

@misc{usda_cdl,
  author       = {{U.S. Department of Agriculture, National Agricultural Statistics Service}},
  title        = {Cropland Data Layer},
  year         = 2024,
  howpublished = {\url{https://nassgeodata.gmu.edu/CropScape/}},
}

@misc{aafc_aci,
  author       = {{Agriculture and Agri-Food Canada}},
  title        = {Annual Crop Inventory},
  year         = 2023,
  howpublished = {\url{https://search.open.canada.ca/openmap/ba2645d5-4458-414d-b196-6303ac06c1c9}},
}

@misc{uk_crome,
  author       = {{UK Rural Payments Agency}},
  title        = {Crop Map of England (CROME)},
  year         = 2023,
  howpublished = {\url{https://environment.data.gov.uk/dataset/a27312b5-d6c9-4710-ad5e-382d727c1b05}},
}

@inproceedings{GSV1,
  title={Combining deep learning and street view imagery to map smallholder crop types},
  author={Soler, Jordi Laguarta and Friedel, Thomas and Wang, Sherrie},
  booktitle={Proceedings of the AAAI Conference on Artificial Intelligence},
  volume={38},
  number={20},
  pages={22202--22212},
  year={2024}
}

@article{median1,
  title={Pre-and within-season crop type classification trained with archival land cover information},
  author={Johnson, David M and Mueller, Richard},
  journal={Remote Sensing of Environment},
  volume={264},
  pages={112576},
  year={2021},
  publisher={Elsevier}
}

@article{median2,
  title={A high-performance and in-season classification system of field-level crop types using time-series Landsat data and a machine learning approach},
  author={Cai, Yaping and Guan, Kaiyu and Peng, Jian and Wang, Shaowen and Seifert, Christopher and Wardlow, Brian and Li, Zhan},
  journal={Remote Sensing of Environment},
  volume={210},
  pages={35--47},
  year={2018},
  publisher={Elsevier}
}

@article{harmonic1,
  title={Phenology-based sample generation for supervised crop type classification},
  author={Belgiu, Mariana and Bijker, Wietske and Csillik, Ovidiu and Stein, Alfred},
  journal={International Journal of Applied Earth Observation and Geoinformation},
  volume={95},
  pages={102264},
  year={2021},
  publisher={Elsevier}
}

@article{harmonic2,
  title={Sentinel-2 cropland mapping using pixel-based and object-based time-weighted dynamic time warping analysis},
  author={Belgiu, Mariana and Csillik, Ovidiu},
  journal={Remote Sensing of Environment},
  volume={204},
  pages={509--523},
  year={2018},
  publisher={Elsevier}
}

@article{tempcnn,
  title={Temporal convolutional neural network for the classification of satellite image time series},
  author={Pelletier, Charlotte and Webb, Geoffrey I and Petitjean, Fran{\c{c}}ois},
  journal={Remote Sensing},
  volume={11},
  number={5},
  pages={523},
  year={2019},
  publisher={MDPI}
}

@article{medianTransfer,
  title={A robust method for mapping soybean by phenological aligning of Sentinel-2 time series},
  author={Huang, Xin and Vrieling, Anton and Dou, Yue and Belgiu, Mariana and Nelson, Andrew},
  journal={ISPRS Journal of Photogrammetry and Remote Sensing},
  volume={218},
  pages={1--18},
  year={2024},
  publisher={Elsevier}
}

@misc{QSC,
  author       = {{Queensland Government}},
  title        = {Queensland Seasonal Crop (QSC) Mapping — Summer and Winter Crop Classifications},
  year         = {2023},
  howpublished = {\url{https://www.qld.gov.au/environment/land/management/mapping/statewide-monitoring/crops}},
}

@misc{LGP,
  author       = {{Flemish Department of Agriculture and Fisheries}},
  title        = {Landbouwgebruikspercelen (LGP) — Annual Crop Parcel Declarations in Belgium},
  year         = {2023},
  howpublished = {\url{https://landbouwcijfers.vlaanderen.be/open-geodata-landbouwgebruikspercelen}},
}

@misc{CTNC,
  author       = {Li, Qiangzi and Wang, Hongyan and Zhang, Yuan and Du, Xin and Dong, Yong and Shen, Yunqi and others},
  title        = {Mapping crop type in Northeast China during 2017--2023 (Including code for classification)},
  year         = {2024},
  howpublished = {figshare. Dataset},
  doi          = {10.6084/m9.figshare.25346038.v1},
  url          = {https://doi.org/10.6084/m9.figshare.25346038.v1},
}

@misc{RPG,
  author       = {{Agence de Services et de Paiement}},
  title        = {Registre Parcellaire Graphique (RPG) — Annual Agricultural Parcel Database of France},
  year         = {2023},
  howpublished = {\url{https://geoservices.ign.fr/rpg}},
}

@misc{BRP,
  author       = {{Netherlands Enterprise Agency}},
  title        = {Basisregistratie Gewaspercelen (BRP) — Annual Agricultural Parcel and Crop Type Register of the Netherlands},
  year         = {2023},
  howpublished = {\url{https://service.pdok.nl/rvo/brpgewaspercelen/atom/v1_0/basisregistratie_gewaspercelen_brp.xml}},
}

@inproceedings{NDVI,
  author    = {Rouse, J. W. and Haas, R. H. and Schell, J. A. and Deering, D. W.},
  title     = {Monitoring vegetation systems in the Great Plains with ERTS},
  booktitle = {Third Earth Resources Technology Satellite-1 Symposium},
  series    = {NASA Special Publication},
  volume    = {SP-351},
  pages     = {309--317},
  year      = {1973},
  address   = {Washington, D.C.},
  publisher = {NASA}
}

@inproceedings{MagnitudeWarping1,
  title={Generating synthetic time series to augment sparse datasets},
  author={Forestier, Germain and Petitjean, Fran{\c{c}}ois and Dau, Hoang Anh and Webb, Geoffrey I and Keogh, Eamonn},
  booktitle={IEEE International Conference on Data Mining},
  pages={865--870},
  year={2017},
  organization={IEEE}
}

@inproceedings{MagnitudeWarping2,
  title={Time series data augmentation for neural networks by time warping with a discriminative teacher},
  author={Iwana, Brian Kenji and Uchida, Seiichi},
  booktitle={International Conference on Pattern Recognition},
  pages={3558--3565},
  year={2021},
  organization={IEEE}
}

@article{MagnitudeWarping3,
  title={An empirical survey of data augmentation for time series classification with neural networks},
  author={Iwana, Brian Kenji and Uchida, Seiichi},
  journal={Plos One},
  volume={16},
  number={7},
  pages={e0254841},
  year={2021},
  publisher={Public Library of Science San Francisco, CA USA}
}

@inproceedings{SpatialDropout,
  title     = {Efficient object localization using Convolutional Networks},
  author    = {Tompson, Jonathan and Goroshin, Ross and Jain, Arjun and LeCun, Yann and Bregler, Christoph},
  booktitle = {Proceedings of the IEEE Conference on Computer Vision and Pattern Recognition},
  year      = {2015},
  pages     = {648--656}
}

@inproceedings{ResNet,
  title={Deep residual learning for image recognition},
  author={He, Kaiming and Zhang, Xiangyu and Ren, Shaoqing and Sun, Jian},
  booktitle={Proceedings of the IEEE Conference on Computer Vision and Pattern Recognition},
  pages={770--778},
  year={2016}
}

@inproceedings{EfficientNet,
  title={Efficientnetv2: Smaller models and faster training},
  author={Tan, Mingxing and Le, Quoc},
  booktitle={International Conference on Machine Learning},
  pages={10096--10106},
  year={2021},
  organization={PMLR}
}

@inproceedings{ConvNeXt,
  title={A convnet for the 2020s},
  author={Liu, Zhuang and Mao, Hanzi and Wu, Chao-Yuan and Feichtenhofer, Christoph and Darrell, Trevor and Xie, Saining},
  booktitle={Proceedings of the IEEE/CVF Conference on Computer Vision and Pattern Recognition},
  pages={11976--11986},
  year={2022}
}

@inproceedings{cams,
  title={Learning deep features for discriminative localization},
  author={Zhou, Bolei and Khosla, Aditya and Lapedriza, Agata and Oliva, Aude and Torralba, Antonio},
  booktitle={Proceedings of the IEEE Conference on Computer Vision and Pattern Recognition},
  pages={2921--2929},
  year={2016}
}

@article{presto,
  title={Lightweight, pre-trained transformers for remote sensing timeseries},
  author={Tseng, Gabriel and Cartuyvels, Ruben and Zvonkov, Ivan and Purohit, Mirali and Rolnick, David and Kerner, Hannah},
  journal={arXiv preprint arXiv:2304.14065},
  year={2023}
}

@misc{cropharvest,
  title={Cropharvest: A global dataset for crop-type classification},
  author={Tseng, Gabriel and Zvonkov, Ivan and Nakalembe, Catherine Lilian and Kerner, Hannah},
  note={Conference on Neural Information Processing Systems Datasets and Benchmarks Track},
  year={2021}
}

@article{worldcereal,
  title={WorldCereal: a dynamic open-source system for global-scale, seasonal, and reproducible crop and irrigation mapping},
  author={Van Tricht, Kristof and Degerickx, Jeroen and Gilliams, Sven and Zanaga, Daniele and Battude, Marjorie and Grosu, Alex and Brombacher, Joost and Lesiv, Myroslava and Bayas, Juan Carlos Laso and Karanam, Santosh and others},
  journal={Earth System Science Data},
  volume={15},
  number={12},
  pages={5491--5515},
  year={2023},
  publisher={Copernicus Publications G{\"o}ttingen, Germany}
}

@article{queensland,
  title={Detecting the annual areal extent of sugarcane crops in Queensland, Australia},
  author={Pringle, Matthew J},
  journal={Remote Sensing Applications: Society and Environment},
  volume={22},
  pages={100496},
  year={2021},
  publisher={Elsevier}
}

@article{GCVI,
  title={Remote estimation of canopy chlorophyll content in crops},
  author={Gitelson, Anatoly A and Vi{\~n}a, Andr{\'e}s and Ciganda, Ver{\'o}nica and Rundquist, Donald C and Arkebauer, Timothy J},
  journal={Geophysical Research Letters},
  volume={32},
  number={8},
  year={2005},
  publisher={Wiley Online Library}
}

@inproceedings{africa1,
  title={Digital agriculture to fulfil the shortage of horticultural data and achieve food security in sub-Saharan Africa},
  author={Sarron, Julien and Beillouin, Damien and Huat, Jo{\"e}l and Koffi, Jean-Mathias and Diatta, Jeanne and Mal{\'e}zieux, {\'E} and Faye, Emile},
  booktitle={All Africa Horticultural Congress: Transformative Innovations in Horticulture},
  pages={239--246},
  year={2021}
}

@article{africa2,
  title={HarvestStat Africa--harmonized subnational crop statistics for sub-Saharan Africa},
  author={Lee, Donghoon and Anderson, Weston and Chen, Xuan and Davenport, Frank and Shukla, Shraddhanand and Sahajpal, Ritvik and Budde, Michael and Rowland, James and Verdin, Jim and You, Liangzhi and others},
  journal={Scientific Data},
  volume={12},
  number={1},
  pages={690},
  year={2025},
  publisher={Nature Publishing Group UK London}
}

@article{weakly1,
  title={Weakly supervised deep learning for segmentation of remote sensing imagery},
  author={Wang, Sherrie and Chen, William and Xie, Sang Michael and Azzari, George and Lobell, David B},
  journal={Remote Sensing},
  volume={12},
  number={2},
  pages={207},
  year={2020},
  publisher={MDPI}
}

@article{weakly2,
  title={Global high categorical resolution land cover mapping via weak supervision},
  author={Tong, Xin-Yi and Dong, Runmin and Zhu, Xiao Xiang},
  journal={ISPRS Journal of Photogrammetry and Remote Sensing},
  volume={220},
  pages={535--549},
  year={2025},
  publisher={Elsevier}
}

@article{t-SNE,
  title={Visualizing data using t-SNE},
  author={Maaten, Laurens van der and Hinton, Geoffrey},
  journal={Journal of Machine Learning Research},
  volume={9},
  number={Nov},
  pages={2579--2605},
  year={2008}
}

@inproceedings{labeltransfer,
  title={Semi-supervised learning using gaussian fields and harmonic functions},
  author={Zhu, Xiaojin and Ghahramani, Zoubin and Lafferty, John D},
  booktitle={Proceedings of the International Conference on Machine Learning},
  pages={912--919},
  year={2003}
}

@article{weakconstraints,
  title={Posterior regularization for structured latent variable models},
  author={Ganchev, Kuzman and Gra{\c{c}}a, Joao and Gillenwater, Jennifer and Taskar, Ben},
  journal={The Journal of Machine Learning Research},
  volume={11},
  pages={2001--2049},
  year={2010},
  publisher={JMLR. org}
}

@misc{WorldCover,
  author    = {Zanaga, D. and Van De Kerchove, R. and Daems, D. and others},
  title     = {ESA WorldCover 10 m 2021 v200},
  year      = {2022},
  publisher = {Zenodo},
  doi       = {10.5281/zenodo.7254221},
  url       = {https://doi.org/10.5281/zenodo.7254221}
}

@article{1DCNN1,
  title={Demonstration of large area land cover classification with a one dimensional convolutional neural network applied to single pixel temporal metric percentiles},
  author={Zhang, Hankui K and Roy, David P and Luo, Dong},
  journal={Remote Sensing of Environment},
  volume={295},
  pages={113653},
  year={2023},
  publisher={Elsevier}
}

@article{1DCNN2,
  title={Evaluating deep learning methods applied to Landsat time series subsequences to detect and classify boreal forest disturbances events: The challenge of partial and progressive disturbances},
  author={Perbet, Pauline and Guindon, Luc and Cote, Jean-Francois and Beland, Martin},
  journal={Remote Sensing of Environment},
  volume={306},
  pages={114107},
  year={2024},
  publisher={Elsevier}
}

@article{Alphaearth,
  title={Alphaearth foundations: An embedding field model for accurate and efficient global mapping from sparse label data},
  author={Brown, Christopher F and Kazmierski, Michal R and Pasquarella, Valerie J and Rucklidge, William J and Samsikova, Masha and Zhang, Chenhui and Shelhamer, Evan and Lahera, Estefania and Wiles, Olivia and Ilyushchenko, Simon and others},
  journal={arXiv preprint arXiv:2507.22291},
  year={2025}
}

@article{harvesting,
  title={Harvesting AlphaEarth: Benchmarking the Geospatial Foundation Model for Agricultural Downstream Tasks},
  author={Ma, Yuchi and Shen, Yawen and Swatantran, Anu and Lobell, David B},
  journal={arXiv preprint arXiv:2601.00857},
  year={2025}
}

@article{Temporally,
  title={Temporally transferable crop mapping with temporal encoding and deep learning augmentations},
  author={Pham, Vu-Dong and Tetteh, Gideon and Thiel, Fabian and Erasmi, Stefan and Schwieder, Marcel and Frantz, David and van der Linden, Sebastian},
  journal={International Journal of Applied Earth Observation and Geoinformation},
  volume={129},
  pages={103867},
  year={2024},
  publisher={Elsevier}
}

@article{TimeMatch,
  title={TimeMatch: Unsupervised cross-region adaptation by temporal shift estimation},
  author={Nyborg, Joachim and Pelletier, Charlotte and Lef{\`e}vre, S{\'e}bastien and Assent, Ira},
  journal={ISPRS Journal of Photogrammetry and Remote Sensing},
  volume={188},
  pages={301--313},
  year={2022},
  publisher={Elsevier}
}

@inproceedings{ANMC,
  title={First large extent and high resolution cropland and crop type map of Argentina},
  author={De Abelleyra, Diego and Veron, S and Banchero, Santiago and Mosciaro, Maria Jesus and Propato, T and Ferraina, Antonella and Taffarel, MC Gomez and Dacunto, Luciana and Franzoni, Agustin and Volante, J},
  booktitle={IEEE Latin American GRSS \& ISPRS Remote Sensing Conference},
  pages={392--396},
  year={2020},
  organization={IEEE}
}

\clearpage
\appendix
\section*{\LARGE Appendix}
\vspace{7mm}

\section{Public Access Links for Reference Crop Type Data}
\label{sec:appendix_links}

\begin{itemize}
    \item[-] Argentina National Map of Crops (ANMC): 

    \url{https://zenodo.org/records/10103323}

\vspace{-2mm}
    \item[-] Queensland Seasonal Crop (QSC): 

    \url{https://www.qld.gov.au/environment/land/management/mapping/statewide-monitoring/crops}

\vspace{-2mm}
    \item[-] Landbouwgebruikspercelen (LGP): 

    \url{https://landbouwcijfers.vlaanderen.be/open-geodata-landbouwgebruikspercelen}

\vspace{-2mm}
    \item[-] Crop Type in Northeast China (CTNC):

    \url{https://figshare.com/articles/dataset/Mapping_crop_type_in_Northeast_China_during_2017-2023_Including_code_for_classification_/25346038}

\vspace{-2mm}
    \item[-] Registre Parcellaire Graphique (RPG): 

    \url{https://geoservices.ign.fr/rpg}

\vspace{-2mm}
    \item[-] Crop Map of England (CROME): 

    \url{https://environment.data.gov.uk/dataset/a27312b5-d6c9-4710-ad5e-382d727c1b05}

\vspace{-2mm}
    \item[-] Basisregistratie Gewaspercelen (BRP): 

    \url{https://service.pdok.nl/rvo/brpgewaspercelen/atom/v1_0/basisregistratie_gewaspercelen_brp.xml}

\vspace{-2mm}
    \item[-] Cropland Data Layer (CDL): 

    \url{https://croplandcros.scinet.usda.gov/}
\end{itemize}

\begin{table*}[htb!]
\caption{RF/MLP comparison for different representations. For harmonic, median, Presto, and AlphaEarth, we evaluate both RF and MLP. In Table~\ref{table:results2}, we report only the best-performing classifier for each representation.}
\vspace{3mm}
\resizebox{\textwidth}{53mm}{
\begin{tabular}{llrrcrrcrr}
\hline
\multirow{10}{*}{\begin{tabular}[c]{@{}l@{}}Cross-\\ Country\end{tabular}}	
& \multirow{2}{*}{Method} & \multicolumn{2}{c}{FRA $\rightarrow$ BEL} & & \multicolumn{2}{c}{FRA $\rightarrow$ NLD} & & \multicolumn{2}{c}{FRA $\rightarrow$ GBR} \\
\cline{3-4}\cline{6-7}\cline{9-10}
&& OA (\%) & mF1 (\%) && OA (\%) & mF1 (\%) && OA (\%) & mF1 (\%) \\
\cline{2-10}
& Harmonic\_RF    & \textbf{80.99 $\pm$ 0.15} & 83.14 $\pm$ 0.12 & & 78.22 $\pm$ 0.24 & 79.83 $\pm$ 0.22 & & \textbf{69.54 $\pm$ 0.20} & \textbf{71.05 $\pm$ 0.17} \\
& Harmonic\_MLP   & 80.86 $\pm$ 0.52 & \textbf{84.77 $\pm$ 0.32} & & \textbf{78.40 $\pm$ 0.81} & \textbf{82.47 $\pm$ 0.55} & & 62.08 $\pm$ 1.52 & 66.05 $\pm$ 1.36 \\
\cline{2-2}
& Median\_RF      & 92.28 $\pm$ 0.09 & 94.56 $\pm$ 0.06 & & 89.18 $\pm$ 0.11 & 90.45 $\pm$ 0.09 & & 89.37 $\pm$ 0.13 & 90.10 $\pm$ 0.09 \\
& Median\_MLP     & \textbf{96.20 $\pm$ 0.71} & \textbf{96.79 $\pm$ 0.50} & & \textbf{93.20 $\pm$ 0.91} & \textbf{93.72 $\pm$ 0.82} & & \textbf{91.84 $\pm$ 0.78} & \textbf{92.07 $\pm$ 0.73} \\
\cline{2-2}
& Presto\_RF      & 96.35 $\pm$ 0.06 & 96.44 $\pm$ 0.06 & & \textbf{91.09 $\pm$ 0.12} & \textbf{91.00 $\pm$ 0.13} & & \textbf{85.62 $\pm$ 0.42} & \textbf{85.89 $\pm$ 0.40} \\
& Presto\_MLP     & \textbf{96.39 $\pm$ 0.35} & \textbf{96.50 $\pm$ 0.33} & & 88.39 $\pm$ 1.41 & 88.16 $\pm$ 1.54 & & 82.11 $\pm$ 3.94 & 82.42 $\pm$ 3.96 \\
\cline{2-2}
& AlphaEarth\_RF  & 96.07 $\pm$ 0.06 & 96.38 $\pm$ 0.06 & & 90.86 $\pm$ 0.26 & 90.75 $\pm$ 0.27 & & 92.77 $\pm$ 0.12 & 92.69 $\pm$ 0.12 \\
& AlphaEarth\_MLP & \textbf{96.99 $\pm$ 0.22} & \textbf{97.63 $\pm$ 0.19} & & \textbf{92.72 $\pm$ 0.65} & \textbf{92.81 $\pm$ 0.69} & & \textbf{93.73 $\pm$ 0.56} & \textbf{93.68 $\pm$ 0.56} \\
\hline

\multirow{10}{*}{\begin{tabular}[c]{@{}l@{}}Cross-\\ Continent\end{tabular}}	
& \multirow{2}{*}{Method} & \multicolumn{2}{c}{USA $\rightarrow$ FRA} & & \multicolumn{2}{c}{FRA $\rightarrow$ CHN} & & \multicolumn{2}{c}{FRA $\rightarrow$ USA} \\
\cline{3-4}\cline{6-7}\cline{9-10}
&& OA (\%) & mF1 (\%) && OA (\%) & mF1 (\%) && OA (\%) & mF1 (\%) \\
\cline{2-10}
& Harmonic\_RF    & \textbf{53.35 $\pm$ 0.18} & 50.28 $\pm$ 0.21 & & 56.57 $\pm$ 0.75 & 52.34 $\pm$ 0.98 & & 45.10 $\pm$ 0.09 & 38.71 $\pm$ 0.12 \\
& Harmonic\_MLP   & 52.20 $\pm$ 0.74 & \textbf{54.71 $\pm$ 0.78} & & \textbf{74.60 $\pm$ 0.89} & \textbf{75.07 $\pm$ 0.92} & & \textbf{53.15 $\pm$ 0.66} & \textbf{50.22 $\pm$ 1.05} \\
\cline{2-2}
& Median\_RF      & 75.24 $\pm$ 0.15 & 74.93 $\pm$ 0.19 & & \textbf{62.65 $\pm$ 1.75} & \textbf{62.13 $\pm$ 2.79} & & \textbf{57.14 $\pm$ 0.08} & 50.19 $\pm$ 0.11 \\
& Median\_MLP     & \textbf{81.42 $\pm$ 1.86} & \textbf{82.41 $\pm$ 1.93} & & 44.64 $\pm$ 5.30 & 38.04 $\pm$ 5.61 & & 56.43 $\pm$ 1.30 & \textbf{51.24 $\pm$ 1.75} \\
\cline{2-2}
& Presto\_RF      & \textbf{62.24 $\pm$ 0.53} & 53.56 $\pm$ 0.88 & & \textbf{51.94 $\pm$ 0.25} & \textbf{34.84 $\pm$ 0.51} & & \textbf{43.35 $\pm$ 0.32} & \textbf{27.11 $\pm$ 0.50} \\
& Presto\_MLP     & 61.55 $\pm$ 3.35 & \textbf{60.59 $\pm$ 4.52} & & 31.33 $\pm$ 2.78 & 16.20 $\pm$ 4.09 & & 38.02 $\pm$ 2.79 & 26.85 $\pm$ 3.06 \\
\cline{2-2}
& AlphaEarth\_RF  & 45.69 $\pm$ 1.36 & 45.47 $\pm$ 1.27 & & 35.77 $\pm$ 0.91 & 22.82 $\pm$ 1.18 & & 68.14 $\pm$ 2.16 & 67.77 $\pm$ 2.53 \\
& AlphaEarth\_MLP & \textbf{61.74 $\pm$ 2.69} & \textbf{61.20 $\pm$ 3.79} & & \textbf{39.64 $\pm$ 6.54} & \textbf{29.46 $\pm$ 9.43} & & \textbf{79.53 $\pm$ 2.53} & \textbf{79.24 $\pm$ 3.69} \\
\hline

\multirow{10}{*}{\begin{tabular}[c]{@{}l@{}}Cross-\\ Hemisphere\end{tabular}}	
& \multirow{2}{*}{Method} & \multicolumn{2}{c}{USA $\rightarrow$ AUS} & & \multicolumn{2}{c}{USA $\rightarrow$ ARG} & & \multicolumn{2}{c}{FRA $\rightarrow$ ARG} \\
\cline{3-4}\cline{6-7}\cline{9-10}
&& OA (\%) & mF1 (\%) && OA (\%) & mF1 (\%) && OA (\%) & mF1 (\%) \\
\cline{2-10}
& Harmonic\_RF    & 61.12 $\pm$ 0.15  & 58.05 $\pm$ 0.16 & & 44.90 $\pm$ 0.23 & 43.64 $\pm$ 0.25 & & 39.33 $\pm$ 0.20 & 33.69 $\pm$ 0.25 \\
& Harmonic\_MLP   & \textbf{64.52 $\pm$ 1.12}  & \textbf{64.78 $\pm$ 1.01} & & \textbf{54.48 $\pm$ 0.70} & \textbf{52.92 $\pm$ 0.62} & & \textbf{45.66 $\pm$ 0.80} & \textbf{44.41 $\pm$ 1.00} \\
\cline{2-2}
& Median\_RF      & 69.84 $\pm$ 0.40  & 66.14 $\pm$ 0.51 & & \textbf{68.89 $\pm$ 0.17} & \textbf{70.13 $\pm$ 0.13} & & 46.62 $\pm$ 0.26 & 38.72 $\pm$ 0.34 \\
& Median\_MLP     & \textbf{77.01 $\pm$ 1.93}  & \textbf{77.58 $\pm$ 1.72} & & 67.52 $\pm$ 1.31 & 67.94 $\pm$ 1.43 & & \textbf{53.24 $\pm$ 2.58} & \textbf{49.02 $\pm$ 2.63} \\
\cline{2-2}
& Presto\_RF      & \textbf{45.13 $\pm$ 1.11}  & \textbf{30.19 $\pm$ 1.19} & & \textbf{33.36 $\pm$ 3.15} & \textbf{21.20 $\pm$ 4.13} & & \textbf{43.40 $\pm$ 1.63} & \textbf{26.98 $\pm$ 0.69} \\
& Presto\_MLP     & 33.80 $\pm$ 10.49 & 22.90 $\pm$ 9.74 & & 27.33 $\pm$ 9.23 & 15.53 $\pm$ 7.59 & & 33.77 $\pm$ 0.00 & 12.63 $\pm$ 0.01 \\
\cline{2-2}
& AlphaEarth\_RF  & \textbf{47.57 $\pm$ 0.03}  & 21.88 $\pm$ 0.08 & & 24.82 $\pm$ 0.14 &  7.25 $\pm$ 0.35 & & \textbf{33.76 $\pm$ 0.00} & 12.68 $\pm$ 0.08 \\
& AlphaEarth\_MLP & 46.31 $\pm$ 10.16 & \textbf{30.90 $\pm$ 8.86} & & \textbf{29.25 $\pm$ 6.46} & \textbf{18.29 $\pm$ 5.98} & & 28.07 $\pm$ 8.44 & \textbf{13.62 $\pm$ 4.36} \\
\hline
\end{tabular}}
\label{table:appendix_classifier}
\end{table*}

\section{Classifiers for Representations}  
\label{sec:appendix_classifier}
Table~\ref{table:appendix_classifier} reports the full RF/MLP comparison for harmonic, median, Presto, and AlphaEarth representations. These results are used to select the best-performing classifier for each feature type and support the summary in Table~\ref{table:results2}. The preferred classifier varies across transfer settings, but overall the strongest choices are Harmonic\_MLP, Median\_MLP, Presto\_RF, and AlphaEarth\_MLP.

\section{Sensitivity Analysis} 
\label{sec:appendix_sensitivity}
\begin{figure*}[t]
\centering
\includegraphics[width=1\textwidth]
{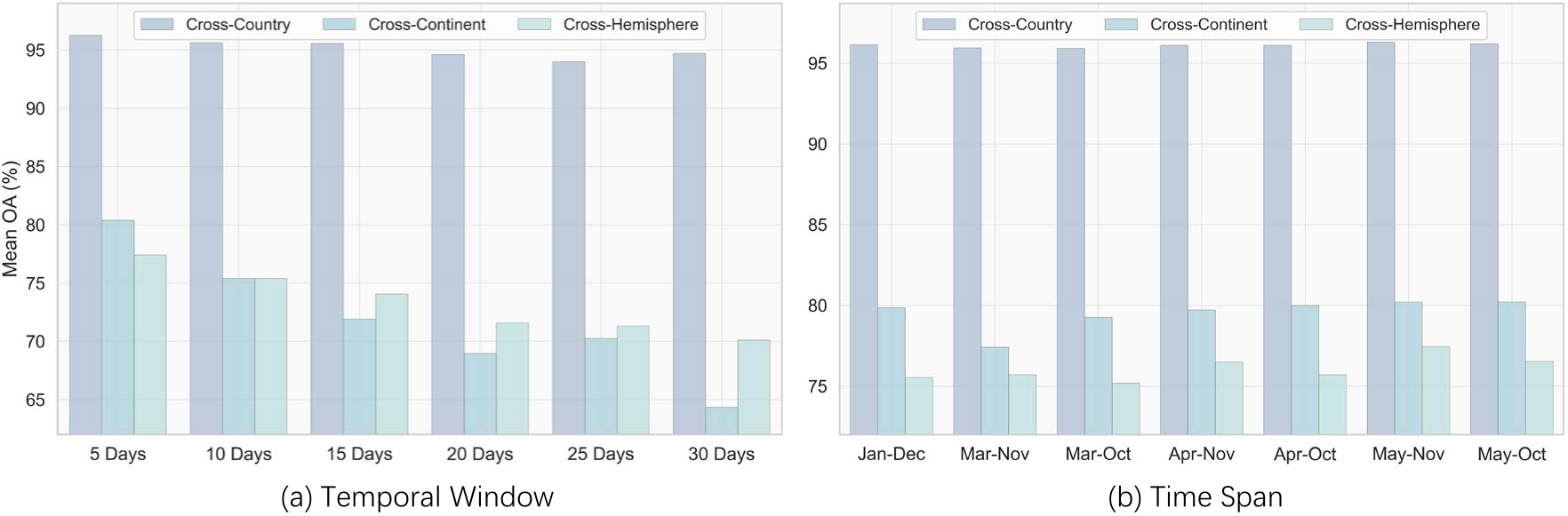}
\caption{Impact of (a) different temporal windows and (b) different time spans on mean OA (\%) across cross-country, cross-continent, and cross-hemisphere settings.}
\label{figure:appendixtime}
\end{figure*}

We conduct a sensitivity analysis to investigate how temporal window and time span affect model performance across transfer settings. As shown in Fig.~\ref{figure:appendixtime}, results are reported as mean OA averaged over all transfer directions within each setting.

In terms of temporal window, shorter windows (e.g., 5-day intervals) consistently lead to better classification results. This suggests that longer intervals (e.g., 15 or 30 days) may compress multiple growth stages into a single observation, causing critical signals to be smoothed out. 

Regarding time span, the impact differs by transfer type. For cross-country transfer, the choice of time span has little effect because FRA, BEL, NLD, and GBR are geographically close and have similar climate, crop calendars, and phenology.

\begin{figure*}[t!]
\centering
\includegraphics[width=1\textwidth]
{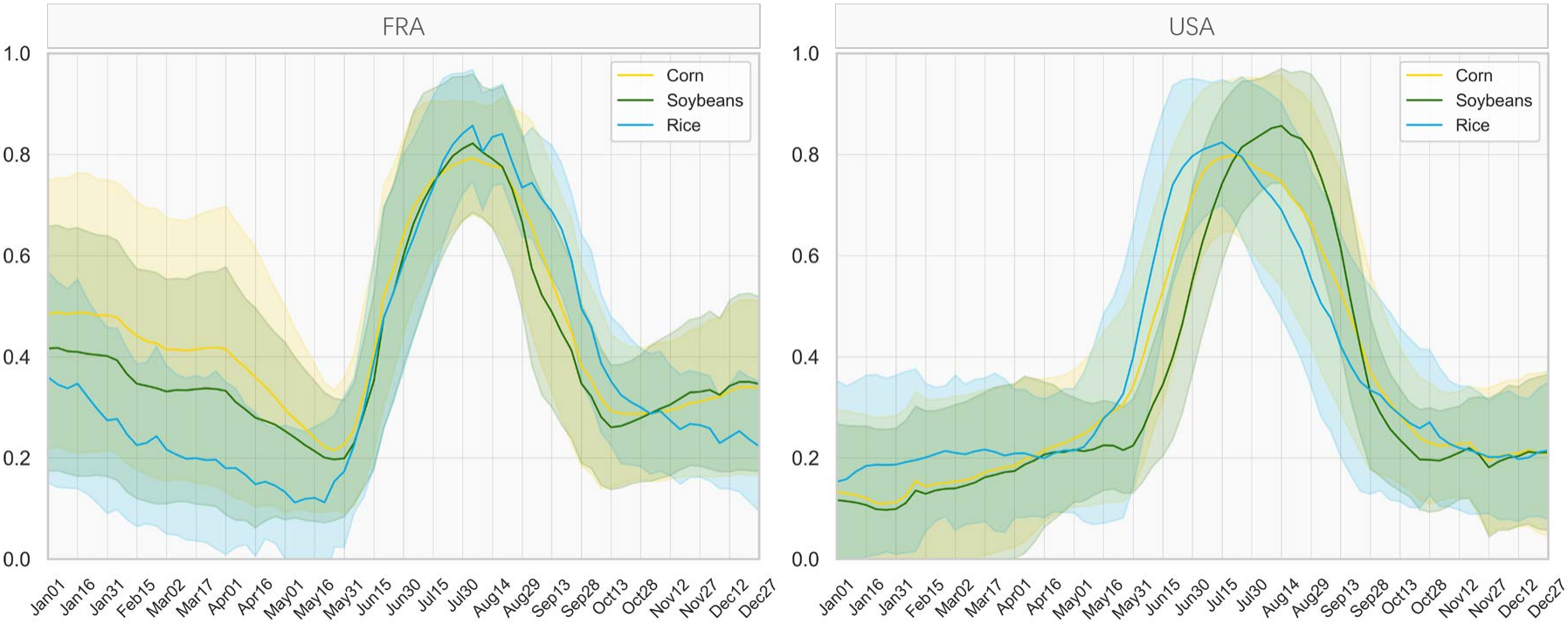}
\caption{NDVI time series for corn, soybeans, and rice in FRA and USA over a full year. In France, NDVI values are already high from January to April, indicating the presence of vegetation during the early months of the year. This may be due to early-sown crops, winter crops, or crop rotation practices.}
\label{figure:appendixndvi}
\end{figure*}

In contrast, cross-continent results are more sensitive to time span. The best performance is achieved with May--October, whereas including early-year months (January---March) degrades performance. To analyze this, we visualize the average NDVI of corn, soybeans, and rice in FRA and USA (Fig.~\ref{figure:appendixndvi}). In FRA, we observe significant NDVI activity during January to March, particularly for corn and soybeans. This is likely due to crop rotation practices (e.g., planting rapeseed in the off-season), which elevate NDVI values and introduce noisy signals unrelated to the target summer crops. When such signals from non-growing season are included, the model may overfit to these phenological patterns, which do not appear in target regions (e.g., USA, where early-year NDVI remains low). 

For cross-hemisphere transfer, the best performance is achieved with May---November. Moreover, we find that including November consistently improves accuracy regardless of the start month. This is because crops such as wheat, sugarcane, and cotton often have longer growth cycles. Considering the trade-offs across different settings, we select May---November as the time span used in our main experiments.

To further refine our understanding of the data augmentation design, we conduct a sensitivity analysis on the key parameters in time shift, time scale, and magnitude warping. These experiments are carried out under four challenging transfer settings: FRA $\rightarrow$ CHN, FRA $\rightarrow$ USA, FRA $\rightarrow$ ARG and USA $\rightarrow$ AUS. Results are reported as the mean OA averaged across these transfer tasks.

\begin{figure*}[t]
\centering
\includegraphics[width=1\textwidth]
{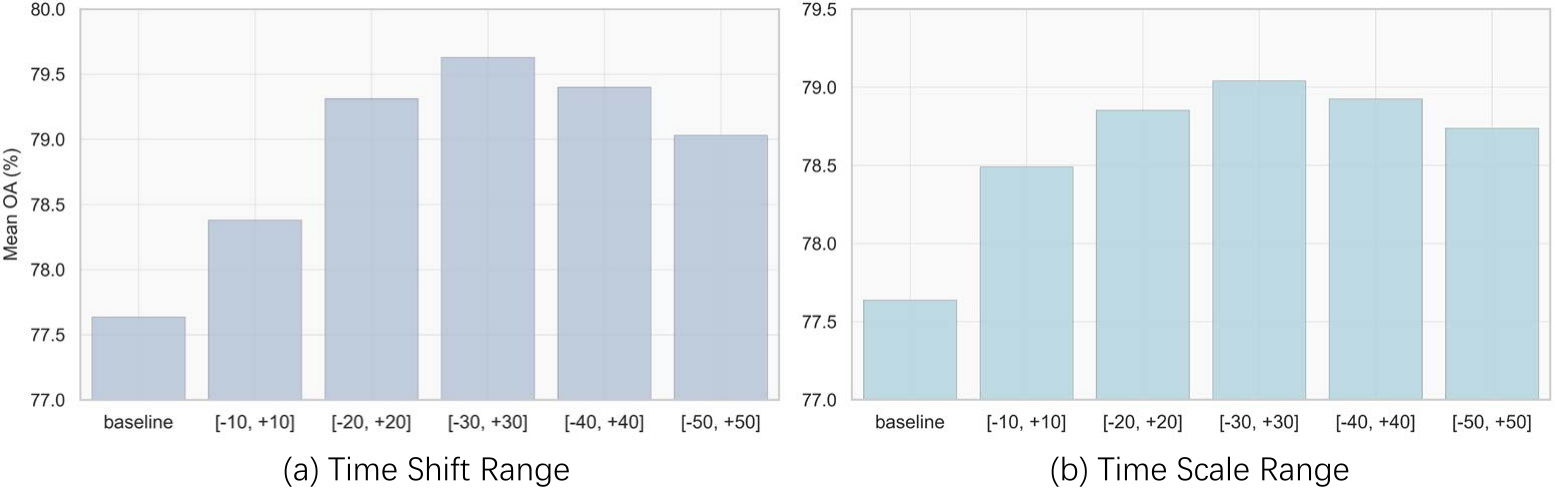}
\caption{Sensitivity analysis of parameters for data augmentation: (a) impact of time shift range; (b) impact of time scale range. Only one transformation (either time shift or time scale) is applied in each experiment. Mean OA (\%) values are averaged over the FRA $\rightarrow$ CHN, FRA $\rightarrow$ USA, FRA $\rightarrow$ ARG and USA $\rightarrow$ AUS transfer settings.}
\label{figure:appendixrange}
\end{figure*}

Fig.~\ref{figure:appendixrange} presents the results when applying time shift or time scale alone. For time shift, performance peaks at a shift range of $[-30, +30]$. When the range becomes larger, accuracy starts to decline. Note that our base time span is fixed to May--November; when a shifted window extends beyond the current year, we clip it to the year boundary and compute median composites on the remaining valid dates. Therefore, overly large shifts may effectively truncate the seasonal profile, reducing the amount of informative growing-season observations.

A similar trend is observed for time scale. Allowing moderate expansion or contraction of the season improves performance, with the best result achieved at $[-30, +30]$. More aggressive scaling leads to slightly worse results, suggesting that overly strong stretching/compression can distort crop development patterns.

\begin{figure*}[t!]
\centering
\includegraphics[width=1\textwidth]
{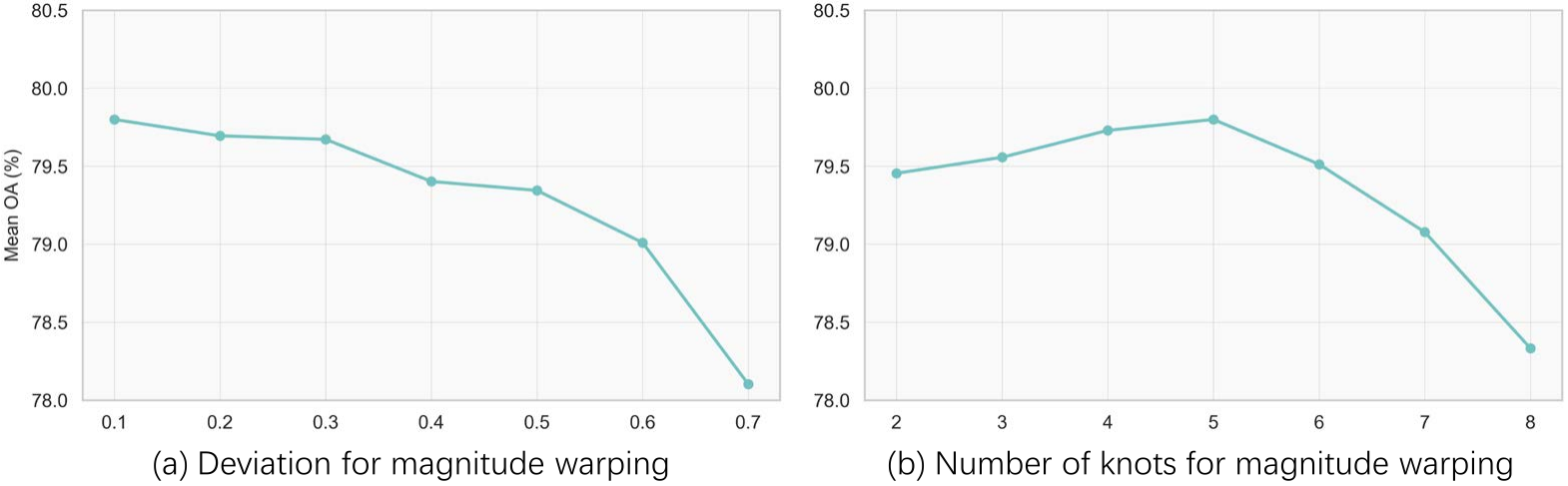}
\caption{Sensitivity analysis of (a) Gaussian deviation and (b) the number of spline knots for magnitude warping. Only magnitude warping is applied in this set of experiments. Mean OA (\%) values are averaged over the FRA $\rightarrow$ CHN, FRA $\rightarrow$ USA, FRA $\rightarrow$ ARG and USA $\rightarrow$ AUS transfer settings.}
\label{figure:appendixwarping}
\end{figure*}

Fig.~\ref{figure:appendixwarping} investigates the parameters of magnitude warping. The left panel shows that performance is highest at $\sigma=0.1$ and gradually decreases as $\sigma$ increases, indicating that magnitude warping is most effective when it introduces realistic but not overly strong reflectance perturbations. The right panel shows that performance peaks at $knots=5$ and declines when fewer or more knots are used, suggesting that this setting best balances smoothness and flexibility.

\section{S1 + S2 Feature Combination}
\label{sec:appendix_s1s2}
Sentinel-1 (S1) synthetic aperture radar (SAR) data are also widely used for crop type classification. SAR is insensitive to clouds and illumination conditions and provides complementary information on canopy structure, vegetation water content, and soil moisture. As a result, many studies combine S1 and S2 to improve robustness, especially when optical observations are sparse.

We further examine whether S1 provides additional transferable information under cross-region transfer. For each CropGlobe sample, we extract S1 observations at the same location and use VV and VH as additional input channels. We then fuse S1 and S2 by channel-wise concatenation: the original S2 median feature of size $10 \times t$ becomes $12 \times t$ after appending VV and VH, and the same classifier is trained on the fused input.

Table~\ref{table:appendix_s1s2} reports the results. Overall, S1 + S2 improves performance in several transfer settings, with gains often becoming larger when combined with data augmentation. This suggests that under cross-region shifts, SAR provides complementary structure- and moisture-related cues beyond optical reflectance.

\begin{figure*}[t!]
\centering
\includegraphics[width=1\textwidth]
{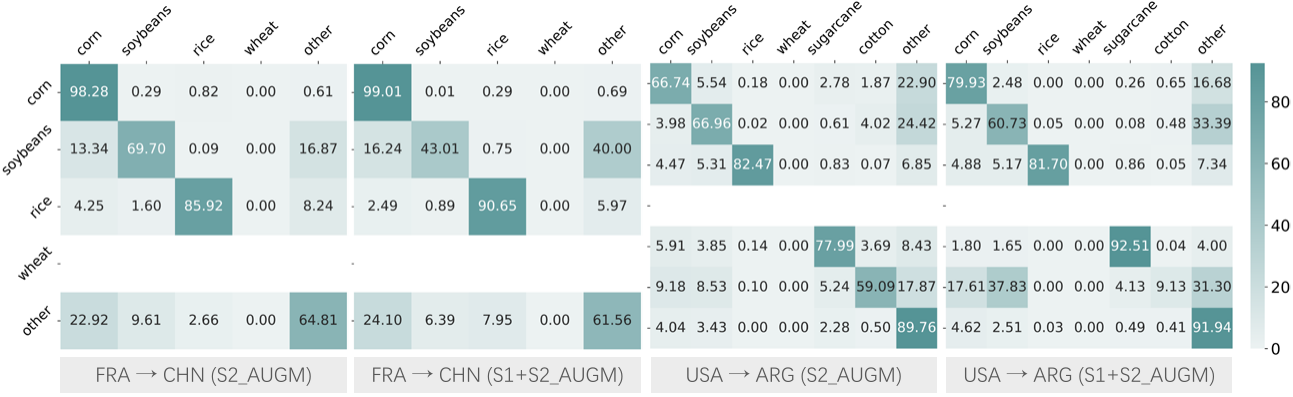}
\caption{Confusion matrices comparing S2 and S1 + S2 inputs with data augmentation under two transfer settings (FRA $\rightarrow$ CHN and USA $\rightarrow$ ARG) from Table~\ref{table:appendix_s1s2}.}
\label{figure:appendixcms12}
\end{figure*}

At the same time, the benefit of S1 is not always consistent. In some settings, OA improves while mF1 decreases (e.g., FRA $\rightarrow$ CHN and USA $\rightarrow$ ARG), indicating gains for dominant classes but worse performance for minority or more heterogeneous categories (Fig.~\ref{figure:appendixcms12}). This may be because SAR responses are affected by soil moisture, tillage, and surface roughness, which vary across regions and introduce additional uncertainty under transfer.

\begin{table*}[t]
\centering
\caption{Experiments on S1 + S2 feature combination. To examine whether SAR can further improve geographic transfer, we download S1 observations at the same coordinates as the CropGlobe samples and use VV and VH as additional channels.}
\vspace{3mm}
\resizebox{0.9\textwidth}{31mm}{
\begin{tabular}{llrrcrrcrr}
\hline
\multirow{6}{*}{\begin{tabular}[c]{@{}l@{}}Cross-\\ Country\end{tabular}}
& \multirow{2}{*}{Data} & \multicolumn{2}{c}{FRA $\rightarrow$ BEL} & & \multicolumn{2}{c}{FRA $\rightarrow$ NLD} & & \multicolumn{2}{c}{FRA $\rightarrow$ GBR} \\
\cline{3-4}\cline{6-7}\cline{9-10}
&& OA (\%) & mF1 (\%) & & OA (\%) & mF1 (\%) & & OA (\%) & mF1 (\%) \\
\cline{2-10}
& S2            & 97.74 $\pm$ 0.55 & 97.99 $\pm$ 0.48 & & 95.85 $\pm$ 0.43 & 95.98 $\pm$ 0.41 & & 95.01 $\pm$ 1.19 & 95.14 $\pm$ 1.19 \\
& S2\_AUGM      & 98.24 $\pm$ 0.18 & 98.39 $\pm$ 0.15 & & 96.15 $\pm$ 0.24 & 96.36 $\pm$ 0.25 & & 95.75 $\pm$ 0.43 & 95.91 $\pm$ 0.46 \\
& S1 + S2       & 98.11 $\pm$ 0.20 & 98.31 $\pm$ 0.20 & & 96.85 $\pm$ 0.15 & 97.05 $\pm$ 0.14 & & 95.54 $\pm$ 0.48 & 95.66 $\pm$ 0.47 \\
& S1 + S2\_AUGM & \textbf{98.29 $\pm$ 0.24} & \textbf{98.44 $\pm$ 0.24} & & \textbf{96.92 $\pm$ 0.24} & \textbf{97.11 $\pm$ 0.21} & & \textbf{96.48 $\pm$ 0.58} & \textbf{96.62 $\pm$ 0.59} \\
\hline

\multirow{6}{*}{\begin{tabular}[c]{@{}l@{}}Cross-\\ Continent\end{tabular}}
& \multirow{2}{*}{Data} & \multicolumn{2}{c}{USA $\rightarrow$ FRA} & & \multicolumn{2}{c}{FRA $\rightarrow$ CHN} & & \multicolumn{2}{c}{FRA $\rightarrow$ USA} \\
\cline{3-4}\cline{6-7}\cline{9-10}
&& OA (\%) & mF1 (\%) & & OA (\%) & mF1 (\%) & & OA (\%) & mF1 (\%) \\
\cline{2-10}
& S2            & 87.00 $\pm$ 0.75 & 89.44 $\pm$ 0.67 & & 79.55 $\pm$ 3.20 & 79.31 $\pm$ 3.44 & & 73.34 $\pm$ 1.95 & 72.60 $\pm$ 2.21 \\
& S2\_AUGM      & 87.76 $\pm$ 1.37 & 90.00 $\pm$ 1.24 & & 82.52 $\pm$ 1.28 & \textbf{82.73 $\pm$ 1.27} & & \textbf{80.37 $\pm$ 1.49} & \textbf{80.68 $\pm$ 1.51} \\
& S1 + S2       & 88.36 $\pm$ 1.34 & 89.88 $\pm$ 1.13 & & 79.19 $\pm$ 5.12 & 71.22 $\pm$ 3.77 & & 76.13 $\pm$ 0.70 & 75.66 $\pm$ 0.89 \\
& S1 + S2\_AUGM & \textbf{89.04 $\pm$ 1.03} & \textbf{90.33 $\pm$ 0.91} & & \textbf{83.42 $\pm$ 0.71} & 75.25 $\pm$ 1.87 & & 80.29 $\pm$ 1.19 & 80.66 $\pm$ 1.25 \\
\hline

\multirow{6}{*}{\begin{tabular}[c]{@{}l@{}}Cross-\\ Hemisphere\end{tabular}}
& \multirow{2}{*}{Data} & \multicolumn{2}{c}{USA $\rightarrow$ AUS} & & \multicolumn{2}{c}{USA $\rightarrow$ ARG} & & \multicolumn{2}{c}{FRA $\rightarrow$ ARG} \\
\cline{3-4}\cline{6-7}\cline{9-10}
&& OA (\%) & mF1 (\%) & & OA (\%) & mF1 (\%) & & OA (\%) & mF1 (\%) \\
\cline{2-10}
& S2            & 86.74 $\pm$ 1.90 & 86.75 $\pm$ 1.87 & & 74.15 $\pm$ 0.42 & 75.18 $\pm$ 0.45 & & 70.92 $\pm$ 2.15 & 70.80 $\pm$ 2.40 \\
& S2\_AUGM      & \textbf{88.45 $\pm$ 0.96} & \textbf{88.38 $\pm$ 1.02} & & 74.84 $\pm$ 0.79 & \textbf{75.63 $\pm$ 0.79} & & 77.03 $\pm$ 1.01 & 77.41 $\pm$ 1.16 \\
& S1 + S2       & 87.09 $\pm$ 0.79 & 86.98 $\pm$ 0.81 & & 79.83 $\pm$ 0.54 & 69.25 $\pm$ 0.62 & & 74.28 $\pm$ 0.96 & 72.39 $\pm$ 1.32 \\
& S1 + S2\_AUGM & 87.86 $\pm$ 0.87 & 87.81 $\pm$ 0.87 & & \textbf{80.88 $\pm$ 0.40} & 70.56 $\pm$ 0.64 & & \textbf{78.87 $\pm$ 0.62} & \textbf{77.72 $\pm$ 0.72} \\
\hline
\end{tabular}}
\label{table:appendix_s1s2}
\end{table*}

In the main text, we focus on S2 because it provides the most widely available and easily reusable time-series input at the global scale. S1 + S2 combination is a valuable direction, and we provide these results in the Appendix as a supplement, leaving a more in-depth multi-sensor study for future work.

\section{Input Settings for Presto}
\label{sec:appendix_presto}
Presto is a reconstruction-based foundation model pretrained to recover masked observations from remote sensing time series. It supports flexible inputs, including single-modality or multi-modality sequences, and can be applied using either full-year data or a subset of months. To examine how its performance depends on the inputs, Table~\ref{table:appendix_presto} compares different settings, varying temporal coverage (all year vs. May--November) and modalities (S2 only vs. S1 + S2).

Restricting the input to May--November generally performs better than using the full year. Although Presto is pretrained with a reconstruction objective, its embedding still depends on the observations provided at inference time. Using only the main growing season may therefore yield representations that are more focused on crop-discriminative patterns. Adding S1 does not consistently improve performance. A possible reason is that the reconstruction objective encourages the model to encode information useful for recovering missing observations rather than features that are most discriminative for cross-region crop classification.

In the main text, for fairness and consistency across methods, we report Presto results using S2 inputs over May--November in Table~\ref{table:results2}.

\begin{table*}[t]
\caption{Presto performance under different input settings and classification heads. We vary temporal coverage: all year (AY), May--November (MN); and input modalities: S2 vs. S1 + S2.}
\vspace{3mm}
\resizebox{\textwidth}{52mm}{
\begin{tabular}{llrrcrrcrr}
\hline
\multirow{9}{*}{\begin{tabular}[c]{@{}l@{}}Cross-\\ Country\end{tabular}}
& \multirow{2}{*}{Setting} & \multicolumn{2}{c}{FRA $\rightarrow$ BEL} & & \multicolumn{2}{c}{FRA $\rightarrow$ NLD} & & \multicolumn{2}{c}{FRA $\rightarrow$ GBR} \\
\cline{3-4}\cline{6-7}\cline{9-10}
&& OA (\%) & mF1 (\%) && OA (\%) & mF1 (\%) && OA (\%) & mF1 (\%) \\
\cline{2-10}
& RF\_S2\_AY       & 94.10 $\pm$ 0.18 & 94.25 $\pm$ 0.18 & & 84.08 $\pm$ 0.27 & 83.48 $\pm$ 0.30 & & 79.86 $\pm$ 0.44 & 80.22 $\pm$ 0.45 \\
& MLP\_S2\_AY      & 95.42 $\pm$ 0.32 & 95.75 $\pm$ 0.29 & & 81.92 $\pm$ 2.48 & 81.67 $\pm$ 2.62 & & 74.03 $\pm$ 3.29 & 73.93 $\pm$ 3.57 \\
& RF\_S2\_MN       & 96.35 $\pm$ 0.06 & 96.44 $\pm$ 0.06 & & 91.09 $\pm$ 0.12 & 91.00 $\pm$ 0.13 & & \textbf{85.62 $\pm$ 0.42} & \textbf{85.89 $\pm$ 0.40} \\
& MLP\_S2\_MN      & 96.39 $\pm$ 0.35 & 96.50 $\pm$ 0.33 & & 88.39 $\pm$ 1.41 & 88.16 $\pm$ 1.54 & & 82.11 $\pm$ 3.94 & 82.42 $\pm$ 3.96 \\
& RF\_S1 + S2\_AY  & 94.14 $\pm$ 0.06 & 94.30 $\pm$ 0.06 & & 84.67 $\pm$ 0.20 & 84.27 $\pm$ 0.22 & & 79.29 $\pm$ 0.36 & 79.57 $\pm$ 0.37 \\
& MLP\_S1 + S2\_AY & 95.46 $\pm$ 0.38 & 95.81 $\pm$ 0.34 & & 78.92 $\pm$ 2.24 & 78.58 $\pm$ 2.46 & & 74.05 $\pm$ 3.36 & 73.96 $\pm$ 3.57 \\
& RF\_S1 + S2\_MN  & 96.59 $\pm$ 0.08 & 96.67 $\pm$ 0.08 & & \textbf{91.45 $\pm$ 0.15} & \textbf{91.38 $\pm$ 0.15} & & 84.74 $\pm$ 0.34 & 85.05 $\pm$ 0.33 \\
& MLP\_S1 + S2\_MN & \textbf{96.71 $\pm$ 0.27} & \textbf{96.85 $\pm$ 0.23} & & 87.46 $\pm$ 1.54 & 87.19 $\pm$ 1.54 & & 79.40 $\pm$ 3.18 & 79.69 $\pm$ 3.34 \\
\hline

\multirow{9}{*}{\begin{tabular}[c]{@{}l@{}}Cross-\\ Continent\end{tabular}}
& \multirow{2}{*}{Setting} & \multicolumn{2}{c}{USA $\rightarrow$ FRA} & & \multicolumn{2}{c}{FRA $\rightarrow$ CHN} & & \multicolumn{2}{c}{FRA $\rightarrow$ USA} \\
\cline{3-4}\cline{6-7}\cline{9-10}
&& OA (\%) & mF1 (\%) && OA (\%) & mF1 (\%) && OA (\%) & mF1 (\%) \\
\cline{2-10}
& RF\_S2\_AY       & 58.15 $\pm$ 0.58 & 48.77 $\pm$ 0.70 & & 35.54 $\pm$ 0.21 & 21.90 $\pm$ 0.22 & & 38.99 $\pm$ 0.23 & 25.62 $\pm$ 0.34 \\
& MLP\_S2\_AY      & 63.14 $\pm$ 4.31 & \textbf{61.33 $\pm$ 4.06} & & 31.81 $\pm$ 2.43 & 17.21 $\pm$ 3.86 & & 38.92 $\pm$ 3.96 & 30.28 $\pm$ 5.06 \\
& RF\_S2\_MN       & 62.24 $\pm$ 0.53 & 53.56 $\pm$ 0.88 & & 51.94 $\pm$ 0.25 & 34.84 $\pm$ 0.51 & & 43.35 $\pm$ 0.32 & 27.11 $\pm$ 0.50 \\
& MLP\_S2\_MN      & 61.55 $\pm$ 3.35 & 60.59 $\pm$ 4.52 & & 31.33 $\pm$ 2.78 & 16.20 $\pm$ 4.09 & & 38.02 $\pm$ 2.79 & 26.85 $\pm$ 3.06 \\
& RF\_S1 + S2\_AY  & 60.65 $\pm$ 0.69 & 52.87 $\pm$ 1.02 & & 44.41 $\pm$ 2.03 & 26.75 $\pm$ 1.10 & & 40.06 $\pm$ 0.21 & 26.61 $\pm$ 0.51 \\
& MLP\_S1 + S2\_AY & 59.78 $\pm$ 5.50 & 58.92 $\pm$ 4.53 & & 30.73 $\pm$ 4.04 & 20.91 $\pm$ 4.46 & & 39.22 $\pm$ 3.38 & \textbf{30.63 $\pm$ 4.50} \\
& RF\_S1 + S2\_MN  & \textbf{65.69 $\pm$ 0.53} & 59.42 $\pm$ 0.94 & & \textbf{59.88 $\pm$ 0.25} & \textbf{35.40 $\pm$ 0.35} & & \textbf{43.83 $\pm$ 0.31} & 28.20 $\pm$ 0.61 \\
& MLP\_S1 + S2\_MN & 59.87 $\pm$ 4.51 & 58.40 $\pm$ 5.29 & & 34.75 $\pm$ 4.55 & 24.49 $\pm$ 4.20 & & 40.20 $\pm$ 2.95 & 29.87 $\pm$ 3.92 \\
\hline

\multirow{9}{*}{\begin{tabular}[c]{@{}l@{}}Cross-\\ Hemisphere\end{tabular}}
& \multirow{2}{*}{Setting} & \multicolumn{2}{c}{USA $\rightarrow$ AUS} & & \multicolumn{2}{c}{USA $\rightarrow$ ARG} & & \multicolumn{2}{c}{FRA $\rightarrow$ ARG} \\
\cline{3-4}\cline{6-7}\cline{9-10}
&& OA (\%) & mF1 (\%) && OA (\%) & mF1 (\%) && OA (\%) & mF1 (\%) \\
\cline{2-10}
& RF\_S2\_AY       & \textbf{52.31 $\pm$ 1.26}  & \textbf{39.94 $\pm$ 1.63}  & & 37.30 $\pm$ 1.73  & \textbf{26.24 $\pm$ 2.16} & & 23.30 $\pm$ 1.26 & 11.86 $\pm$ 1.63 \\
& MLP\_S2\_AY      & 46.70 $\pm$ 12.78 & 37.83 $\pm$ 14.27 & & 34.04 $\pm$ 6.51  & 20.51 $\pm$ 4.57 & & 33.76 $\pm$ 0.00 & 12.62 $\pm$ 0.00 \\
& RF\_S2\_MN       & 45.13 $\pm$ 1.11  & 30.19 $\pm$ 1.19  & & 33.36 $\pm$ 3.15  & 21.20 $\pm$ 4.13 & & 43.40 $\pm$ 1.63 & \textbf{26.98 $\pm$ 0.69} \\
& MLP\_S2\_MN      & 33.80 $\pm$ 10.49 & 22.90 $\pm$ 9.74  & & 27.33 $\pm$ 9.23  & 15.53 $\pm$ 7.59 & & 33.77 $\pm$ 0.00 & 12.63 $\pm$ 0.01 \\
& RF\_S1 + S2\_AY  & 51.58 $\pm$ 1.23  & 39.24 $\pm$ 2.29  & & 48.12 $\pm$ 0.85  & 25.92 $\pm$ 1.88 & & 20.96 $\pm$ 1.73 & 11.47 $\pm$ 1.86 \\
& MLP\_S1 + S2\_AY & 47.38 $\pm$ 11.40 & 36.71 $\pm$ 10.43 & & 38.67 $\pm$ 1.42  & 14.07 $\pm$ 4.18 & & 39.86 $\pm$ 0.00 & 14.25 $\pm$ 0.00 \\
& RF\_S1 + S2\_MN  & 46.57 $\pm$ 1.08  & 32.89 $\pm$ 0.97  & & \textbf{48.42 $\pm$ 1.57}  & 25.41 $\pm$ 1.93 & & \textbf{44.05 $\pm$ 1.76} & 24.26 $\pm$ 2.12 \\
& MLP\_S1 + S2\_MN & 35.84 $\pm$ 11.30 & 25.01 $\pm$ 10.94 & & 28.71 $\pm$ 14.02 & 13.19 $\pm$ 5.87 & & 39.86 $\pm$ 0.00 & 14.25 $\pm$ 0.00 \\
\hline
\end{tabular}}
\label{table:appendix_presto}
\end{table*}

\section{Cross-Temporal Transfer}
\label{sec:appendix_crosstime}
Updating crop maps across years is important in practice for agricultural monitoring, annual statistics, and year-to-year comparisons. At the same time, cross-year crop classification is challenging: even within the same country, inter-annual changes in weather and management can shift crop calendars, growth trajectories, and background conditions, creating temporal distribution shifts.

\begin{table*}[t]
\centering
\caption{Sample distribution across years for the USA data used in the cross-temporal transfer experiments (2017, 2020, and 2023).}
\vspace{3mm}
\resizebox{0.75\textwidth}{9mm}{
\begin{tabular}{lcrcrcrcrrcrcrcr}
\hline
\textbf{Year} & & \textbf{Corn} & & \textbf{Soybeans} & & \textbf{Rice} & & \textbf{Wheat} & \textbf{Sugarcane} & & \textbf{Cotton} & & \textbf{Other} & & \textbf{Total} \\
\hline
USA\_2017 & & 3,391  & & 2,624  & & 4,279  & & 5,705  & 2,358 & & 3,909  & & 9,157  & & \textbf{31,423}  \\
USA\_2020 & & 4,165  & & 2,827  & & 4,287  & & 5,764  & 2,273 & & 3,347  & & 8,316  & & \textbf{30,979}  \\
USA\_2023 & & 22,224 & & 17,093 & & 17,613 & & 23,226 & 9,897 & & 14,799 & & 28,427 & & \textbf{133,279} \\
\hline
\textbf{Total} & & \textbf{29,780} & & \textbf{22,544} & & \textbf{26,179} & & \textbf{34,695} & \textbf{14,528} & & \textbf{22,055} & & \textbf{45,900} & & \textbf{195,681} \\
\hline
\end{tabular}}
\label{table:appendix_usa_data}
\end{table*}

\begin{figure*}[t]
\centering
\includegraphics[width=1\textwidth]
{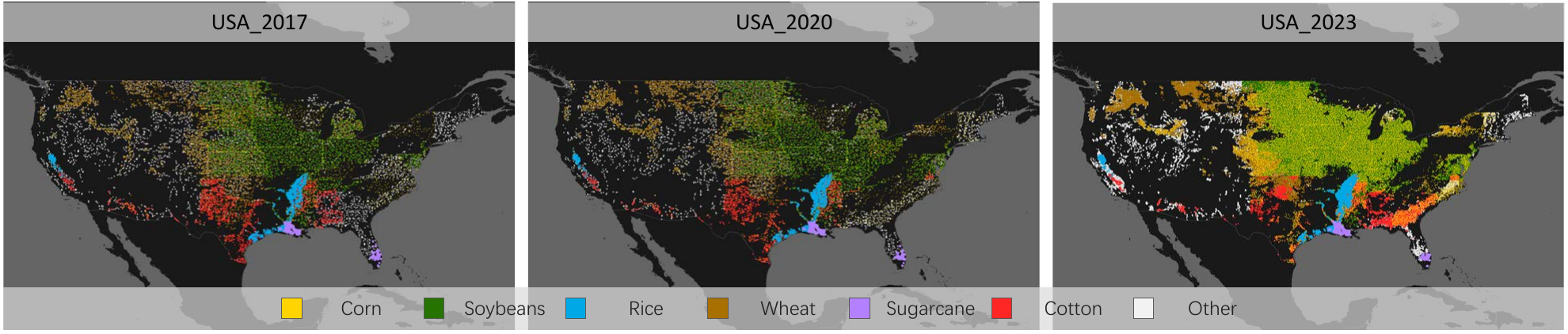}
\caption{Spatial distribution of USA samples used for cross-temporal transfer (USA\_2017, USA\_2020, and USA\_2023).}
\label{figure:appendixusa}
\end{figure*}

Motivated by this, we conduct cross-temporal transfer experiments in the USA: models are trained on USA\_2023 and directly transferred to earlier years (USA\_2017 and USA\_2020) without using target-year labels. Table~\ref{table:appendix_usa_data} summarizes the sample distribution for each year, and Fig.~\ref{figure:appendixusa} visualizes the corresponding spatial coverage.

\begin{figure*}[t]
\centering
\includegraphics[width=1\textwidth]
{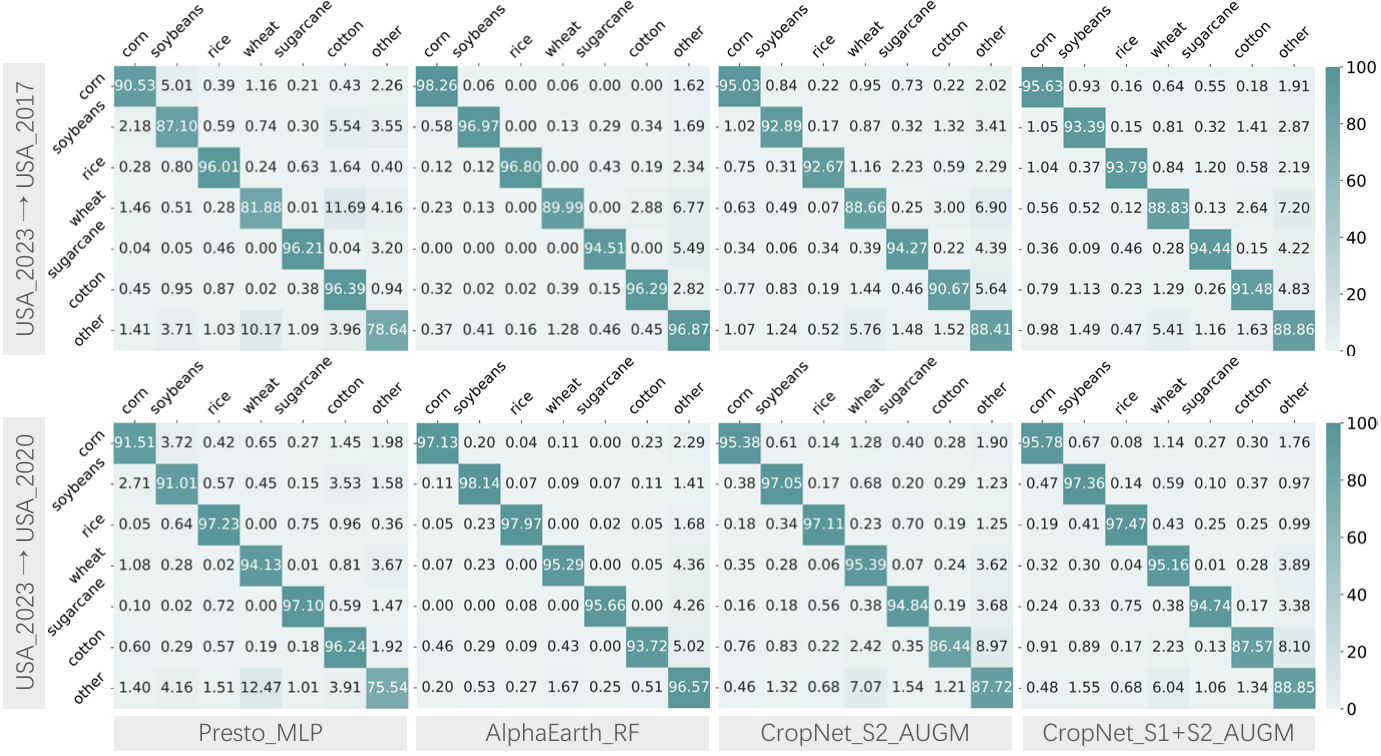}
\caption{Confusion matrices for cross-temporal transfer within the USA from Table~\ref{table:appendix_usa_result}. Models are trained on USA\_2023 and evaluated on USA\_2017 and USA\_2020 without using target-year labels.}
\label{figure:appendixcmusa}
\end{figure*}

Table~\ref{table:appendix_usa_result} reports the results. Across methods, performance on USA\_2023 $\rightarrow$ USA\_2020 is generally higher than on USA\_2023 $\rightarrow$ USA\_2017, which is consistent with the intuition that a larger year gap tends to introduce stronger distribution shifts. To understand how this shift affects different classes, we show confusion matrices in Fig.~\ref{figure:appendixcmusa}.

A notable result is that AlphaEarth performs particularly well in this within-country, cross-year setting, with both RF and MLP reaching nearly 96\% in OA and mF1. Compared with time-series features that rely more directly on within-season trajectories, the annual AlphaEarth embedding provides a more stable year-scale representation. Because it aggregates multi-source information and is designed to be globally consistent, it may be less sensitive to year-specific fluctuations such as weather-driven shifts in temporal curves. As a result, when the main challenge is inter-annual variation within the same region rather than large geographic differences, such a representation can be advantageous. The confusion matrices support this interpretation: AlphaEarth shows fewer concentrated errors and a cleaner diagonal, indicating more stable class separability across years.

Even so, CropNet remains highly competitive in this cross-year setting. With only 133k training samples from USA\_2023 and a lightweight architecture requiring far less computation, CropNet clearly outperforms Presto and achieves strong performance relative to AlphaEarth, a much larger foundation model trained on massive data. This result highlights the strong potential of spectral--temporal representations for cross-temporal crop classification.

\begin{table*}[t]
\centering
\caption{Cross-temporal transfer results within the USA. Models are trained on USA\_2023 and evaluated on USA\_2017 and USA\_2020 without using target-year labels.}
\vspace{3mm}
\resizebox{0.75\textwidth}{15mm}{
\begin{tabular}{lrrcrr}
\hline
\multirow{2}{*}{Method} & \multicolumn{2}{c}{USA\_2023 $\rightarrow$ USA\_2017} & & \multicolumn{2}{c}{USA\_2023 $\rightarrow$ USA\_2020} \\
\cline{2-3}\cline{5-6}
& OA (\%) & mF1 (\%) && OA (\%) & mF1 (\%) \\
\hline
Presto\_RF             & 85.39 $\pm$ 0.08 & 85.66 $\pm$ 0.09 & & 87.68 $\pm$ 0.06 & 88.31 $\pm$ 0.06 \\
Presto\_MLP            & 87.11 $\pm$ 0.33 & 88.00 $\pm$ 0.27 & & 89.38 $\pm$ 0.15 & 90.37 $\pm$ 0.16 \\
AlphaEarth\_RF         & \textbf{95.52 $\pm$ 0.04} & \textbf{96.03 $\pm$ 0.05} & & \textbf{96.37 $\pm$ 0.03} & \textbf{96.81 $\pm$ 0.03} \\
AlphaEarth\_MLP        & 95.11 $\pm$ 0.15 & 95.50 $\pm$ 0.17 & & 96.34 $\pm$ 0.13 & 96.74 $\pm$ 0.12 \\
CropNet\_S2\_AUGM      & 90.85 $\pm$ 0.21 & 91.51 $\pm$ 0.23 & & 92.71 $\pm$ 0.24 & 93.34 $\pm$ 0.22 \\
CropNet\_S1 + S2\_AUGM & 91.38 $\pm$ 0.15 & 92.06 $\pm$ 0.15 & & 93.22 $\pm$ 0.24 & 93.82 $\pm$ 0.23 \\
\hline
\end{tabular}}
\label{table:appendix_usa_result}
\end{table*}

\end{document}